\pdfoutput=1

\documentclass[11pt]{article}

\usepackage[final]{acl}

\usepackage{times}
\usepackage{latexsym}
\usepackage{amssymb}
\usepackage[T1]{fontenc}

\usepackage[utf8]{inputenc}
\usepackage{newunicodechar}
\newunicodechar{β}{\beta}
\usepackage{makecell}
\usepackage{tabularx}
\usepackage{algpseudocode}
\usepackage{array}
\usepackage{algorithm}
\usepackage{microtype}
\usepackage{inconsolata}
\usepackage{graphicx}
\newunicodechar{≤}{\ensuremath{\leq}}
\usepackage{amsmath}
\usepackage{multirow}
\usepackage{amsfonts}
\usepackage{booktabs}
\usepackage{float}
\usepackage{subcaption}
\usepackage{xcolor}
\usepackage[many]{tcolorbox}
\definecolor{grey}{HTML}{f5f5f5}
\definecolor{d_grey}{HTML}{bac8d3}

\newtcolorbox{boxH}{
    colback = grey, 
    colframe = d_grey, 
    boxrule = 0pt, 
    leftrule = 6pt 
}

\title{StateSwap: Probing Support–Elimination Hidden States in Multiple-Choice Questions}

\author{
 \textbf{Chao Gao\textsuperscript{1}},
 \textbf{Haijiang Liu\textsuperscript{2,3}},
 \textbf{Qiyuan Li\textsuperscript{2,3}},
 \textbf{Caicai Guo\textsuperscript{2,3}},
\\
 \textbf{Frank van Harmelen\textsuperscript{1}},
 \textbf{Jinguang Gu\textsuperscript{2,3}}
\\
\\
{\small \textsuperscript{1}Department of Computer Science, Vrije Universiteit Amsterdam, Amsterdam, The Netherlands}\\
{\small \textsuperscript{2}School of Computer Science and Technology, Wuhan University of Science and Technology, Wuhan, China}\\
{\small \textsuperscript{3}Hubei Province Key Laboratory of Intelligent Information Processing and Real-time Industrial System,} \\
{\small Wuhan University of Science and Technology, Wuhan, China}
\\
 \small{
   \textbf{Correspondence:} \href{mailto:c.gao@vu.nl}{c.gao@vu.nl}, \href{mailto:SIMON@wust.edu.cn}{simon@wust.edu.cn}
 }
}

\begin{document}
\maketitle
\begin{abstract}
Large language models often answer the same multiple-choice question inconsistently when it is posed under support-oriented and elimination-oriented framings.
We investigate whether these discrepancies arise from different internal representations induced by {the two framings}. {We introduce a dual-framing protocol with minimally varied prompts that use either support- or elimination-oriented framing while keeping the evaluation target fixed.} To probe the internal computation, we append an untrained special token, \texttt{[STATE]}, and treat its residual-stream activation as an intervention interface.
Across both models, the two framings induce separable \texttt{[STATE]} activations concentrated in intermediate layers. Swapping these activations between paired prompts systematically changes predictions and improves cross-framing agreement, providing intervention-based evidence that the activations are behaviorally relevant. 
Beyond instance-level substitution, mean-difference steering directions derived from the dual-framing contrast exhibit more bounded layer-wise responses than matched contrastive activation addition directions under the evaluated protocol. Code is available at https://github.com/Cha0Ga0/SWAPSTATE

\end{abstract}

\begin{figure}[t]
\centering
\includegraphics[width=\columnwidth]{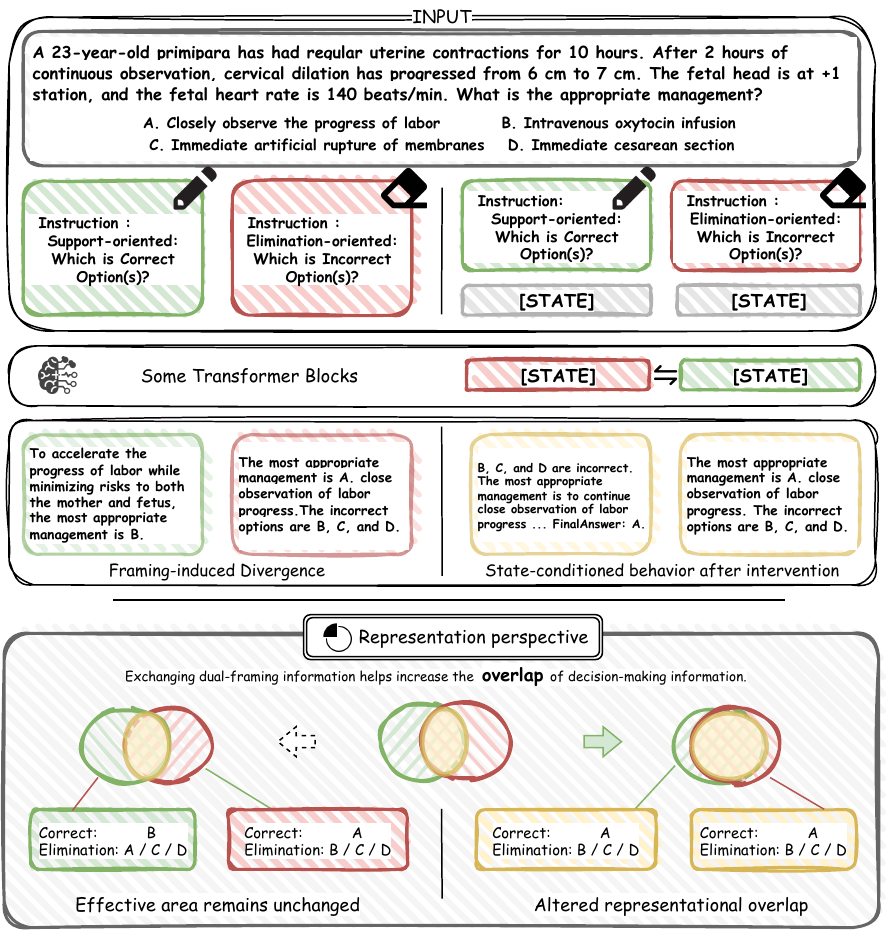}
\caption{
\textbf{Overview of StateSwap.} 
{StateSwap swaps \texttt{[STATE]} activations between paired \textsc{Sup} and \textsc{Elim} prompts while keeping the textual input fixed. The figure shows an idealized agreement-increasing outcome for illustration only; substitution may improve, reduce, or leave unchanged accuracy and agreement. We interpret systematic prediction changes, including but not limited to improvements, as evidence of behavioral relevance under intervention.}
}
\end{figure}
\section{Introduction}

Recent work has shown that the process-of-elimination (PoE) strategy, widely used by humans for solving multiple-choice questions (MCQs), can improve the performance of large language models (LLMs) \cite{ma2023poe}. This approach is motivated by the fact that support-oriented (\textsc{Sup}) prompts (“which option is correct?”) and elimination-oriented (\textsc{Elim}) prompts (“which options are incorrect?”) are logically tied to the same answer key. For example, some studies first prompt the model to identify and eliminate incorrect options before deriving the final answer \cite{zhu2025option, fu2025exclusion, wang2026eliminate, zhao2025elimination}. However, empirical studies also report that elimination-style prompting can instead impair accuracy \cite{balepur2024s}. At first glance, this appears contradictory: if \textsc{Sup} and \textsc{Elim} prompts encode the same task objective and semantic content for decision-making, they should not lead to different predictions. This motivates the following question: do the two framings induce distinguishable internal representations?

{Inspired by recent mechanistic studies \cite{heimersheim2024use, zhang2025finite, lu2025latent}, we investigate prompt framing under controlled inference settings, using deterministic decoding and paired prompt evaluation to isolate representation-level effects from decoding variability. We test whether framing-dependent behavioral differences are accompanied by distinguishable activation patterns in the model's hidden-state space. Separability identifies a representational correlate of prompt framing, and activation substitution evaluates its behavioral relevance.}

{Based on this perspective, we investigate how \textsc{Sup} and \textsc{Elim} prompts shape the internal representations of the model, and whether targeted activation substitution can provide intervention-based evidence on the role of these representations. Specifically, we use a newly registered, untrained special token with no pretrained lexical semantics, \texttt{[STATE]}, to analyze the geometry of the hidden states elicited by the two framings and assess their representational distinguishability. We then introduce activation-level interventions that replace selected internal states while keeping the textual input fixed, allowing us to examine how changes in the induced representation affect the model's output distribution over answer indices. Comparisons with dual-prompt ensemble baselines distinguish activation substitution from simple answer aggregation. Finally, we test whether the intervention remains effective when \texttt{[STATE]} is moved away from its canonical final position.}

The main contributions of this work are threefold: 

\begin{enumerate}
    \item We identify distinguishable \texttt{[STATE]} representations induced by \textsc{Sup} and \textsc{Elim} prompts that encode the same underlying decision problem.
    \item {We show that substituting framing-sensitive \texttt{[STATE]} activations changes model predictions and is associated with higher accuracy and cross-framing agreement in the evaluated settings.}
    \item We find that mean-difference steering directions derived from the two-framing prompt contrast exhibit lower layer-wise variability than matched contrastive activation addition directions under the evaluated protocol.
\end{enumerate}

\begin{figure*}[t]
\centering
\includegraphics[width=\textwidth]{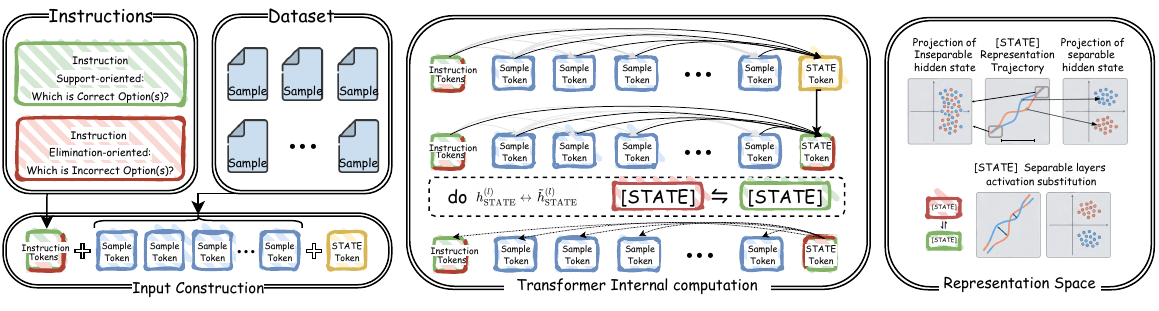}
\caption{
{Paired \textsc{Sup} (green) and \textsc{Elim} (red) prompts place an aligned \texttt{[STATE]} interface (yellow) at the same position. Layer-wise diagnostics localize candidate intervention regions (Section~\ref{sec:method_window}). StateSwap then replaces the full \texttt{[STATE]} activation between the paired framings within the selected layers (Section~\ref{sec:method_swap}).}
}
\label{fig:method_frame}
\end{figure*}

\section{Related Work}

LLM reasoning has been studied from multiple complementary perspectives, including prompting, fine-tuning, symbolic tools, model collaboration, and output verification \citep{he2025solving}. {Sparse-autoencoder analyses characterize activation patterns associated with chain-of-thought reasoning \citep{chen2025does}, while causal tracing and model editing link localized computations to factual predictions \citep{Meng2022ROME}.} Circuit-level studies localize induction-like computations to particular attention heads \citep{Olsson2022InductionHeads}, and activation patching provides a framework for testing the behavioral relevance of localized components \citep{heimersheim2024use}. Synthetic state-tracking experiments further show that transformers can develop localized mechanisms, including late-layer MLP neurons, for representing and updating latent states \citep{zhang2025finite}.

This interventionist view is further supported by recent work on inference-time intervention and activation steering demonstrating that high-level behaviors in LLMs can be influenced by manipulating directions in activation space \cite{li2023inference, turner2024activation, zou2023representation}. Methods such as contrastive activation addition (CAA) \cite{panickssery2023steering}, conditional activation steering \cite{lee2025programming}, and dynamic steering approaches \cite{li2025fairsteer} show that activation shifts derived from dataset-level statistics can systematically alter model outputs, including truthfulness, refusal behavior, and bias-related responses. These results provide strong evidence that internal representations encode behaviorally meaningful information in explorable and manipulable forms. 

A parallel line of work explores interface-based control of internal computation via localized representation slots. In encoder-style models, special tokens such as \texttt{[CLS]} serve as sequence-level interfaces to global representations \citep{devlin2019bert}, while in autoregressive models, parameter-efficient methods such as prompt tuning and prefix tuning learn continuous embeddings that steer model behavior without modifying base parameters \citep{lester-etal-2021-power, li2021prefix, Liu2021PTuningV2}. These approaches highlight the role of localized interfaces as high-leverage access points to latent representations.

Motivated by this line of work, we adopt an intervention-based perspective to investigate how prompt framing shapes internal representations. To operationalize this perspective, we use a fixed, untrained token as a minimal read--write interface to the residual stream. {This setup allows us to examine whether semantically equivalent prompts with different framings give rise to distinguishable decision-related representations for the same input instance and whether these representations can be aligned through instance-level intervention.}

\section{Methodology}
\label{sec:method}

{We investigate whether framing-conditioned \texttt{[STATE]} activations are behaviorally relevant under controlled substitution. Section~\ref{sec:method_interface} defines the aligned inputs and state interface, Section~\ref{sec:method_swap} formalizes cross-framing substitution, and Section~\ref{sec:method_window} describes how the intervention layers are localized; Appendix~\ref{app:tb} summarizes the notation.}

\subsection{Input Construction and State Interface}
\label{sec:method_interface}
{
Consider a causal language model $f_\theta$ with $L$ Transformer blocks. We append a newly registered \texttt{[STATE]} token to each prompt, yielding an input sequence $\mathbf{x}=(x_1,\dots,x_T)$ in which \texttt{[STATE]} occupies the final position $t_S \triangleq T$. Let $\mathbf{h}^{(l)}_t \in \mathbb{R}^d$ denote the residual-stream representation at position $t$ after Transformer block $l$, where $t\in\{1,\dots,T\}$ and $l\in\{1,\dots,L\}$. We define the corresponding \texttt{[STATE]} representation as
\[
\mathbf{s}^{(l)} \triangleq \mathbf{h}^{(l)}_{t_S}.
\]
Although \texttt{[STATE]} has no pretrained lexical semantics and receives no task-specific training, its contextualized representation can aggregate information from the preceding prompt and propagate through later layers to affect subsequent generation. We therefore use $\mathbf{s}^{(l)}$ as a single-token read--write interface for inference-time interventions. Token registration and embedding initialization are described in Appendix~\ref{app:state_token}.
}

For each question $q$, a pair of prompts is constructed, differing only in the {framing instruction}: eliminating incorrect options (\textsc{Elim}) versus supporting correct options (\textsc{Sup}).
Let $I \in \{\textsc{Elim}, \textsc{Sup}\}$ denote the {framing instruction}. Each prompt ends with \texttt{[STATE]}:
\[
\mathbf{x}_I(q)=\mathrm{Concat}(I, q, \texttt{[STATE]}).
\]

To ensure that \texttt{[STATE]} occupies the same token index {in both framings}, instruction segments are padded to the same token length (Appendix~\ref{app:state_align}). 

\subsection{State Substitution and Prediction Sensitivity}
\label{sec:method_swap}

{Let $W\subseteq\{1,\dots,L\}$ denote the contiguous intervention region localized by the diagnostic in Section~\ref{sec:method_window}. For framing $I$, let $\bar I$ denote its complement. Each substitution direction is run independently on $\mathbf{x}_I(q)$, with the cached donor state from $\bar I$ substituted over $W$:}
\[
\mathbf{s}^{(k)}(I,q) \leftarrow
\mathbf{s}^{(k)}_{\mathrm{cache}}(\bar I,q),\quad k\in W.
\]

{ For each $k\in W$, the intervention overwrites only the post-block state $\mathbf{s}^{(k)}(I,q)$ during the prompt-processing pass; all other
residual-stream vectors at the intervention point remain unchanged, although downstream activations may change through residual propagation.
Computation then proceeds normally from the substituted states to standard autoregressive generation. Feature-window slices are used only for
diagnostics and localization of $W$; the intervention itself replaces the full $d$-dimensional state $\mathbf{s}^{(k)}$.
}
\subsection{Locating a Stable Intervention Window}
\label{sec:method_window}

For $\mathbf{s}^{(l)} \in \mathbb{R}^d$, candidate regions are identified by scanning contiguous feature windows of size $w$:
\[
\mathbf{s}^{(l)}_{[j:j+w)} \in \mathbb{R}^{w}, \quad j \in \{0,\dots,d-w\}.
\]

For each layer $l$ and window $(j,w)$, a paired Cohen’s $d$ statistic is computed from the paired question-wise \textsc{Elim}--\textsc{Sup} differences (Appendix~\ref{app:cohen_d}). Two scalarizations are used.

\paragraph{Random-direction diagnostic}
Once per run, a single random unit vector $\mathbf{u}\in\mathbb{R}^d$ is sampled and projected onto the corresponding window slice:
\[
\phi^{\mathrm{rand}}_{j,w}(\mathbf{z})=\mathbf{u}_{[j:j+w)}^\top \mathbf{z},
\]
yielding $d^{\mathrm{rand}}_l(j,w)$.

\paragraph{Mean-difference diagnostic}
For each $(l,j,w)$, we form the mean-difference direction
\[
\mathbf{v}_{l}(j,w)=\frac{1}{N}\sum_{i=1}^N
\left(\mathbf{s}^{(l)}_{\textsc{Elim},i,[j:j+w)}-\mathbf{s}^{(l)}_{\textsc{Sup},i,[j:j+w)}\right),
\]
stabilize its normalization and project:
\[
\begin{aligned}
\phi^{\mathrm{md}}_{l,j,w}(\mathbf{z})
&=\hat{\mathbf{v}}_{l}(j,w)^\top \mathbf{z},\\
\hat{\mathbf{v}}_{l}(j,w)
&=\frac{\mathbf{v}_{l}(j,w)}
{\max(\|\mathbf{v}_{l}(j,w)\|,\epsilon_v)},
\end{aligned}
\]
where $\epsilon_v>0$ is small, yielding $d^{\mathrm{md}}_l(j,w)$.

Window-level separability scores are summarized into a per-layer intensity for each layer. Layers are ranked by this intensity, and { adjacent selected layers are merged into a contiguous intervention region \(W\);} the aggregation and selection procedure is given in Appendix~\ref{app:region_select}.

\section{Experimental Setup and Evaluation Protocol}

\label{sec:exp}

\subsection{Datasets}
To study {prompt-framing} effects under controlled conditions, we evaluate on two multiple-choice benchmarks:
(i) a reasoning-focused subset of MMLU covering 17 subject categories (MMLU-17) \cite{hendrycks2021measuring}, and
(ii) MedQA-CH, a Chinese medical benchmark \cite{jin2021disease},
which require non-trivial reasoning and allow unambiguous comparison across equivalent formulations.
To avoid data leakage, we use 50 calibration examples from the training split solely to localize a candidate intervention window; these examples do not enter downstream test evaluation, and the selected window is fixed for all reported experiments. After fixing the window, we conduct a post hoc robustness analysis using 1,000 examples and 500 bootstrap resamples. Dataset statistics are provided in Appendix~\ref{app:data_stats}.
All downstream inference and reported evaluations use the official test splits.

\subsection{Models and decoding}
We evaluate on reasoning-focused multiple-choice benchmarks and open-weight LLMs Qwen-2.5-7B-Instruct \cite{yang2024qwen25} and GLM-4-9B \cite{glm2024chatglm} in a zero-shot setting. 
All generations use deterministic decoding (greedy; no sampling) to avoid confounding intervention effects with sampling variance.
We fix the chat template, context length, and decoding hyperparameters across all runs; details are provided in Appendix~\ref{app:hyperparameters}.

\subsection{Prompt structure and \texttt{[STATE]}}
All prompts consist of an instruction segment, the question with options, and a final \texttt{[STATE]} token.
The exact instruction templates are provided in Appendix~\ref{sec:prompt_state_appendix}.
To verify robustness to random \texttt{[STATE]} initialization, we rerun our method on randomly sampled instances with different random seeds while keeping all other settings fixed, and observe negligible variation in the generated responses (Appendix~\ref{app:state_init_robustness}).

\subsection{Inference}
We consider {two framings}:
a \textsc{Sup} instruction, which asks the model to predict the correct option(s), and an \textsc{Elim} instruction, which asks the model to predict the incorrect option(s).
For {each framing}, we evaluate a baseline condition and an intervention condition in which the \texttt{[STATE]} representation is substituted over a selected layer region $W$ using {the other framing}.
All other computations are held fixed. Outputs from {both framings} are parsed into a shared option-set representation, as detailed in Appendix~\ref{app:parsing}.

For reported ensemble results, we use paired deterministic prompt realizations within each framing. Thus, \textsc{Sup} and \textsc{Elim} each consist of two prompt-instantiated runs, yielding two framing-specific voting groups. We use strict-majority voting within each group.

\subsection{Metrics}
We report accuracy (ACC) for {each framing} and cross-framing decision overlap measured by the Jaccard index.
We further analyze correctness to distinguish symmetric and asymmetric outcomes.
Response-level stability is evaluated using Length-Aware Cosine Similarity (LAC) and BERTScore-F1 computed on the full generated responses under deterministic decoding.
Cosine similarity captures global embedding-level alignment, while BERTScore-F1 reflects token-level semantic consistency between responses.
Metric definitions and parsing rules are provided in Appendix~\ref{app:metrics}.

\section{Results}

We evaluate the proposed framework through the following evaluation questions (EQs):

\begin{description}
\item[EQ1 (Representation)]
Do {the two framings} induce separable internal representations?
\item[EQ2 (Effects of State Substitution)]
Does exchanging the final-position \texttt{[STATE]} representation transfer decision-relevant information {between the two framings}?
\item[EQ3 (Mechanism Specificity)]
Are these effects specific to the \texttt{[STATE]} interface rather than generic perturbations?
\end{description}

We additionally conduct a separate extension analysis to examine whether the underlying \textsc{Sup}--\textsc{Elim} contrast remains useful without the \texttt{[STATE]} interface in a population-level steering setting.

\subsection{EQ1: Separable Decision-Related Representations}
\label{sec:exp_internal}

Before evaluating the downstream effects of \texttt{[STATE]} substitution, we examine whether {the two framings} produce systematically different internal representations at the \texttt{[STATE]} interface.
{In this diagnostic step, representation-level separability localizes a plausible layer window $W$ for subsequent interventions.}

\begin{figure}[t]
\centering
\includegraphics[width=\columnwidth]{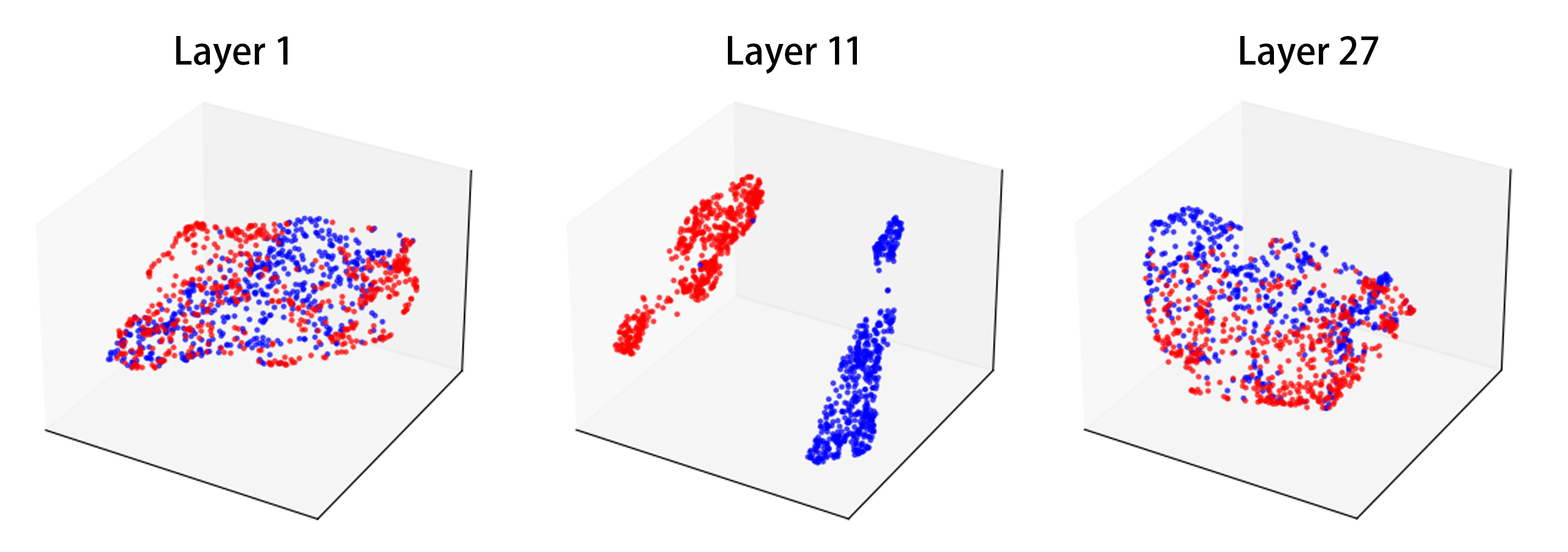}
\caption{
PCA visualization of \texttt{[STATE]} representations for Qwen-2.5-7B at Layers 1, 11, and 27.
Red and blue points denote \textsc{Sup} and \textsc{Elim} prompts, respectively.
The two framings are largely mixed in the early layers, become clearly separated in the intermediate layers, and show weaker separation in the later layers. Additional layer-wise PCA visualizations are provided in Appendix~\ref{app:pca_vis}.
}
\label{fig:pca_state_qwen}
\end{figure}

\paragraph{Separability diagnostics}
Figure~\ref{fig:pca_state_qwen} visualizes the \texttt{[STATE]} representations across representative layers.
The \textsc{Sup} and \textsc{Elim} representations are largely mixed in the early layers, become clearly separated in the intermediate layers, and show weaker separation in the later layers.
This pattern suggests that framing-dependent structure at the \texttt{[STATE]} position is concentrated in a subset of intermediate layers rather than uniformly distributed across depth.
We further quantify this trend with layer-wise separability diagnostics for both Qwen-2.5-7B and GLM-4-9B in Appendix~\ref{app:cross_model_separability}.

\paragraph{Localization of the intervention window $W$}
Based on the intermediate-layer concentration observed in Figure~\ref{fig:pca_state_qwen} and the layer-wise separability diagnostics, we select a compact intervention window $W$ for \texttt{[STATE]} substitution.
For Qwen-2.5-7B, we use layers 11--20 as the intervention window.
For GLM-4-9B, the same separability-based criterion selects layers 21--40; full cross-model diagnostics are reported in Appendix~\ref{app:cross_model_separability}.
These windows are determined from representation-level separability diagnostics rather than downstream task performance.
A post-hoc comparison across candidate Qwen layer ranges, including the magnitude and stability of output-distribution shifts under substitution, is provided in Appendix~\ref{app:window_analysis}. The stability of the localized intermediate-layer region under a larger calibration sample and bootstrap resampling is evaluated in Appendix~\ref{app:window_robustness}.

\paragraph{Random-label control}
To assess whether the observed layer-wise separability could arise from spurious correlations or artifacts of the analysis procedure, we perform a random-label control in which the binary {framing labels} are randomly permuted while all representations and projection settings are held fixed.
This provides a baseline for estimating the level of separability expected under chance alignment.

\begin{figure}[t]
\centering
\includegraphics[width=\columnwidth]{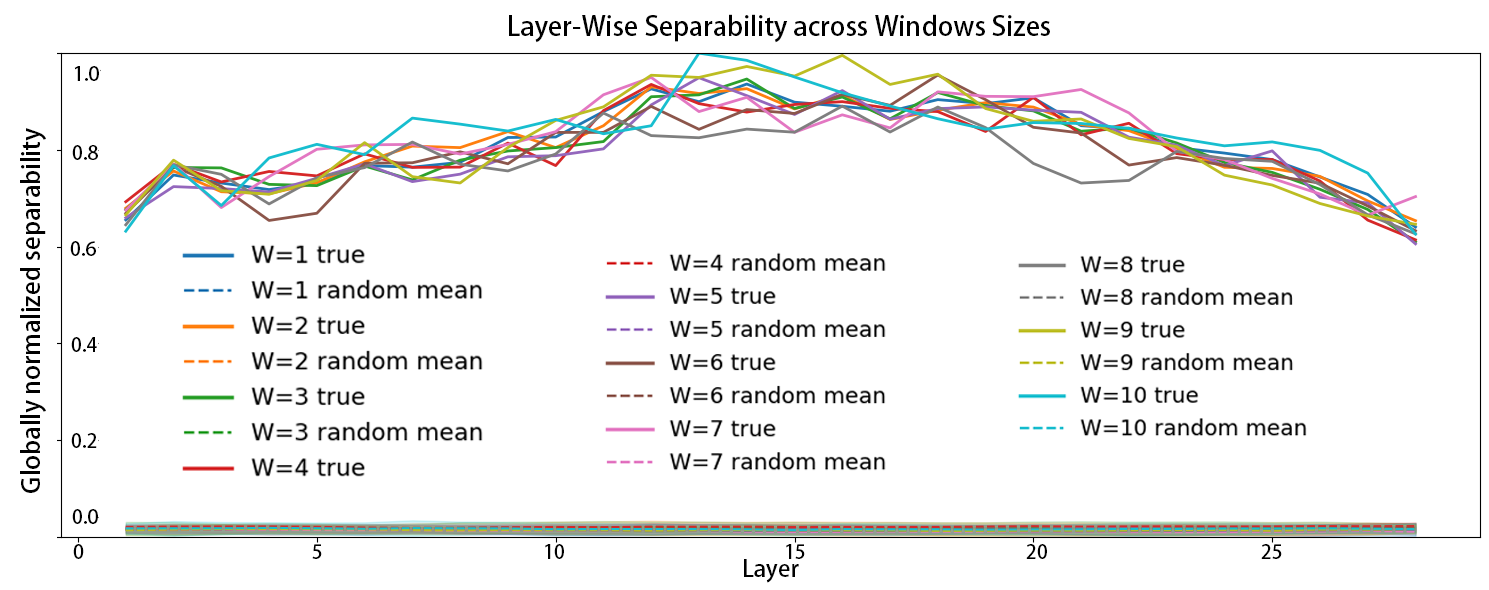}
\caption{Random-label control for layer-wise separability across window sizes.
Solid lines: true labels. Dashed lines: shuffled-label mean across repeated shuffles.
Shaded regions: $\pm 2\sigma$ across shuffles.
{For each layer and feature-window size $w$, the separability score is the Top-$K$ mean of the absolute paired Cohen's $d$ values across feature-window positions. Scores are jointly min--max normalized using the union of the true-label values and the shuffled-label mean $\pm 2\sigma$ bounds across all feature-window sizes.}}
\label{fig:cohen_d_shuffle}
\end{figure}

As shown in Figure~\ref{fig:cohen_d_shuffle}, the separability obtained from 10 random label permutations remains consistently low across layers and window sizes, in contrast to the structured intermediate-layer peak observed under the true labels.
The gap between the two conditions persists across all window sizes considered, indicating that the localization pattern is not an artifact of windowing, random projection, or label imbalance.

\subsection{EQ2: Effects of State Substitution}
\label{sec:behavioral_effects}

\paragraph{Decision-related behavioral effects of \texttt{[STATE]} substitution}

\begin{table}[t]
\centering
\small
\setlength{\tabcolsep}{4pt}
\begin{tabular}{lccc}
\toprule
\textbf{Setting} & \textbf{ACC (\textsc{Sup})} & \textbf{ACC (\textsc{Elim})} & \textbf{Jaccard} \\
\midrule \midrule
\multicolumn{4}{c}{\textbf{Qwen-2.5-7B}} \\ \midrule
MMLU-17 (base) & 65.52 & 65.38 & 75.46\\
MMLU-17 (Sub)  & \textbf{66.53} & \textbf{66.31} & \textbf{76.63}\\ \midrule
MedQA-CH (base) & 77.12 & 74.36 & 78.10\\
MedQA-CH (Sub)  & \textbf{80.26} & \textbf{77.45} & \textbf{81.58}\\ \midrule
\multicolumn{4}{c}{\textbf{GLM-4-9B}} \\ 
\midrule
MMLU-17 (base) & 66.47 & 68.74 & 72.96\\
MMLU-17 (Sub)  & \textbf{69.21} & \textbf{72.36} & \textbf{78.21}\\ \midrule
MedQA-CH (base) & 75.91 & 73.54 & 78.67\\
MedQA-CH (Sub)  & \textbf{76.82} & \textbf{77.92} & \textbf{82.48}\\
\bottomrule
\end{tabular}
\caption{\textbf{Behavioral outcomes under cross-framing [STATE] substitution.} 
{Base denotes inference with the aligned [STATE]-augmented prompt but without activation replacement. Sub denotes cross-framing replacement over the selected layer window \(W\). For ACC (SUP), the recipient run uses a SUP prompt and receives the cached [STATE] activation from its paired ELIM prompt; ACC (ELIM) uses the reverse direction. Each value is computed by within-framing strict-majority voting over two deterministic prompt realizations, with ties counted as incorrect. SUP and ELIM predictions are not combined. Jaccard is the intersection-over-union of the question-index sets answered correctly under SUP and ELIM. All values are percentages; higher values indicate higher accuracy or greater overlap between successful instances.}
All values are percentages; higher is better. Per-task MMLU accuracy is reported in Table~\ref{tab:mmlu_swap_reordered}.}
\label{tab:acc_jaccard_main}
\end{table}

Table~\ref{tab:acc_jaccard_main} examines whether substituting the localized \texttt{[STATE]} representation leads to observable downstream changes in decision behavior.
{Greedy decoding removes sampling randomness from the \textsc{Base}$\rightarrow$\textsc{Sub} comparison.}
Across both models, benchmarks, and {framings}, \texttt{[STATE]} substitution consistently changes the model's prediction outcomes.
In particular, both {framing-specific} accuracy and cross-framing Jaccard overlap increase under substitution.
{These changes provide behavioral evidence that the substituted \texttt{[STATE]} activations affect decision-related computation in addition to exhibiting linear separability.}
More fine-grained accuracy breakdowns and ensemble comparisons are provided in Appendix~\ref{app:ensemble_ablation}.

\paragraph{{Framing} asymmetry under intervention}

\begin{figure}[t]
\centering
    \includegraphics[width=\columnwidth]{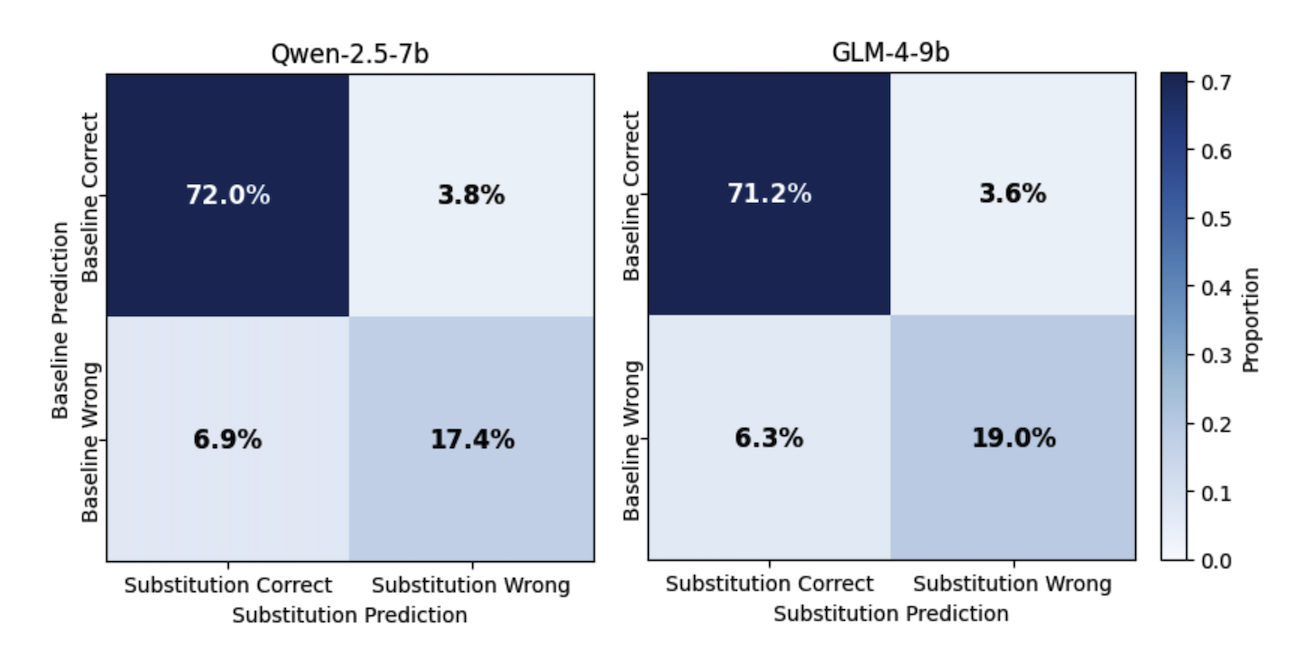}
\caption{
Base-to-substitution accuracy transition matrices.
Each matrix shows the proportion of examples transitioning between baseline and substituted predictions, aggregated over {both framings}.
Rows indicate baseline outcomes (Correct / Wrong), and columns indicate outcomes after \texttt{[STATE]} substitution.
The four cells correspond to Correct$\rightarrow$Correct, Correct$\rightarrow$Wrong, Wrong$\rightarrow$Correct, and Wrong$\rightarrow$Wrong transitions.
Results are shown for Qwen-2.5-7B (left) and GLM-4-9B (right).
}
\label{fig:joint_correctness}
\end{figure}

Figure~\ref{fig:joint_correctness} summarizes the accuracy transitions induced by \texttt{[STATE]} substitution for both models. Each matrix reports how predictions shift between correct and incorrect outcomes when replacing the original decision state with its counterpart from {the other framing}.

{Across both Qwen-2.5-7B and GLM-4-9B, substantial off-diagonal mass shows that substitution systematically alters model behavior. A non-trivial fraction of previously incorrect predictions becomes correct after substitution, while a smaller fraction of correct predictions degrades. The imbalance between these transitions reveals an asymmetric behavioral effect across the two framing directions. Additional analysis in Appendix~\ref{sec:decision_determinacy} shows that \texttt{[STATE]} substitution more often increases than decreases eliminated-incorrect mass, suggesting enhanced error-elimination determinacy.}

\subsection{EQ3: Interface, Content, and Position Specificity}
\label{sec:ablation_exp}

\paragraph{Ablation settings}
We conduct a series of controlled ablations to isolate which aspects of the intervention are responsible for the observed behavioral effects.
For each ablation, we compare the intervened output against the model's original greedy-decoding output without activation intervention.
All interventions are applied within the same layer window $W$ identified in Section~\ref{sec:exp_internal}.
The ablations are organized along three orthogonal dimensions:

\begin{enumerate}
    \item \textbf{Intervention object}
    {We compare StateSwap at \texttt{[STATE]} with a final-content-token control that applies the same swap to the original prompt's final content token without \texttt{[STATE]}. This comparison measures interface specificity.}

    \item \textbf{Substitution content}
    {We vary what replaces the original \texttt{[STATE]} representation, comparing structured cross-framing substitutions with content-agnostic controls. This comparison measures the dependence of model behavior on substitution content.}

    \item \textbf{Interface position}
    {We move \texttt{[STATE]} from its canonical final position to earlier positions relative to the question and option segments while keeping the substitution operation and layer window fixed. This comparison measures position dependence.}

\end{enumerate}

Detailed implementations of each ablation, including zeroing, random replacement, {framing-conditioned} substitution, and alternative interface positions, are provided in Appendix~\ref{app:ablation_details}.

\begin{figure}[t]
\centering
\begin{subfigure}[t]{\columnwidth}
  \centering
    \includegraphics[width=\columnwidth]{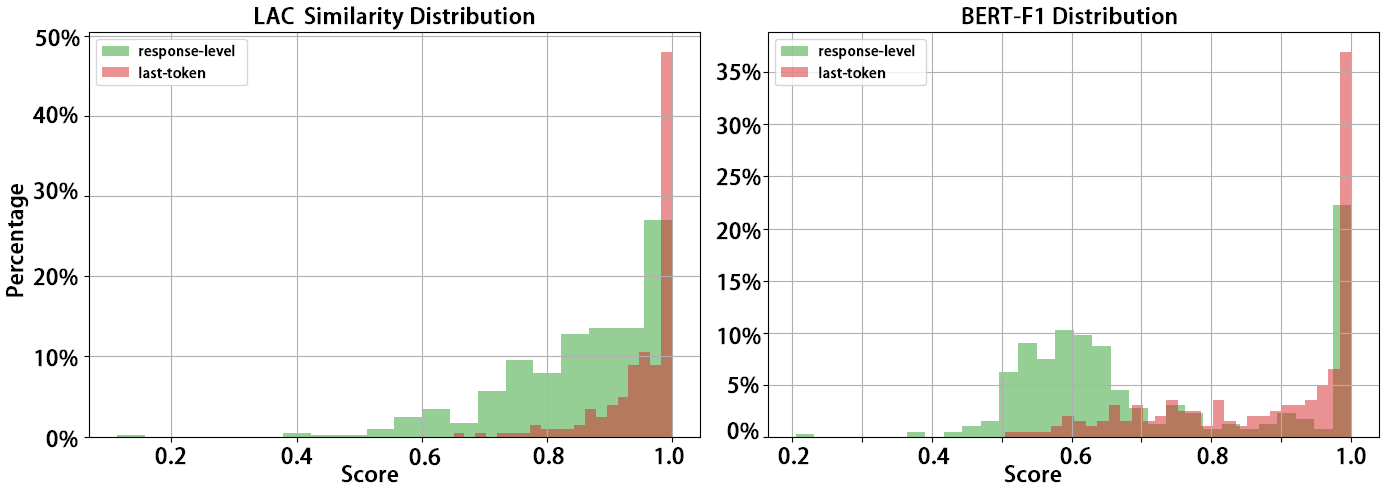}
  \caption{Interface specificity. Substituting \texttt{[STATE]} vs.\ the final content token under the same window $W$ and greedy decoding.}
  \label{fig:dist_a}
\end{subfigure}

\vspace{1mm}

\begin{subfigure}[h]{\columnwidth}
  \centering
    \includegraphics[width=\columnwidth]{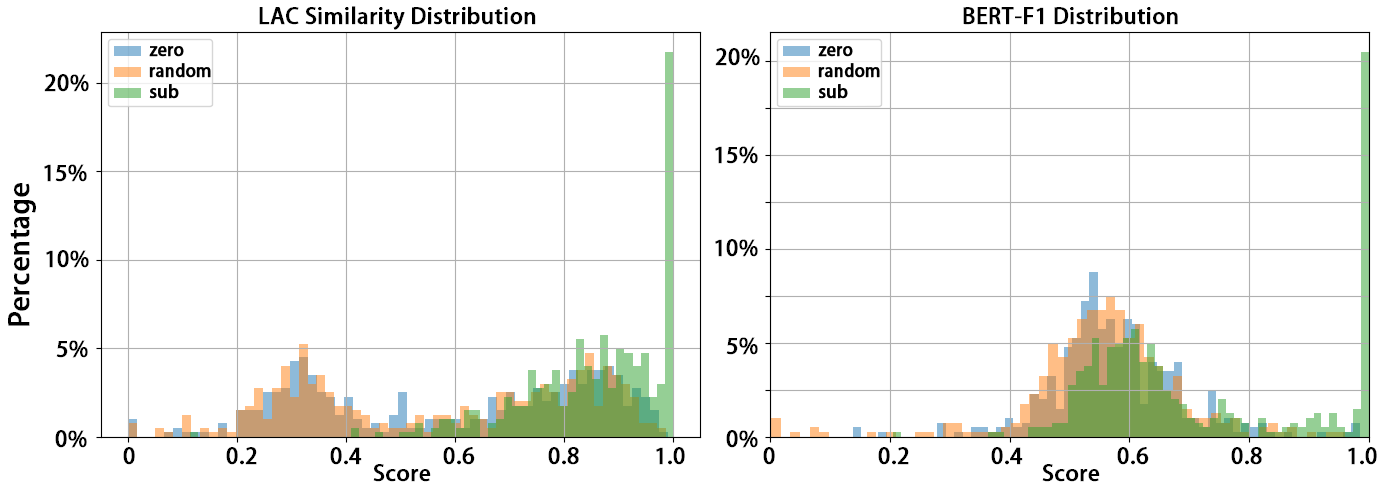}
  \caption{Ablations at \texttt{[STATE]}. Zeroing and random replacement vs.\ {cross-framing substitution}.}
  \label{fig:dist_b}
\end{subfigure}

\caption{{Response-level stability controls. (a) shows that late-position content-token states are substantially less effective as intervention targets than the \texttt{[STATE]} interface. (b) distinguishes unstructured residual injections (zero/random), which degrade stability, from structured cross-framing substitution, which preserves high-similarity regimes.}}
\label{fig:distribution_controls}
\end{figure}

\paragraph{Interface specificity: \texttt{[STATE]} versus final-content-token substitution}
We assess whether the observed effects depend on the intervention location or can be reproduced by modifying late-stage token representations more generally. We apply the same substitution procedure to the original final content token, while keeping the intervention window and decoding configuration fixed.

As shown in Figure~\ref{fig:dist_a}, substitution at the final content token produces outputs that remain close to the baseline, with similarity scores concentrated near 1.0.
Under the tested conditions, the \texttt{[STATE]} interface is therefore more effective than the final-content-token control at altering model behavior.
Quantile summaries for both conditions are reported in Table~\ref{tab:quant_lasttok}.

\paragraph{Effect of substitution content}
We examine how the nature of the substituted representation affects model behavior.
Figure~\ref{fig:dist_b} reports response-level similarity between the baseline output and outputs produced under different intervention types, measured using length-aware cosine similarity (LAC) and BERTScore-F1. Quantile statistics for both metrics are provided in Table~\ref{tab:quant_subst_content}.

Unstructured interventions (e.g., zeroing or random replacement) substantially reduce response similarity, whereas {cross-framing substitution} preserves high similarity across most examples.
{The distinct response regimes associate StateSwap's effects with the content of the substituted activation.}

\paragraph{\texttt{[STATE]} token ablation}
{We also evaluate the effect of inserting the \texttt{[STATE]} token on MedQA-CH accuracy by comparing the original prompt with an otherwise identical prompt that includes the token. Adding \texttt{[STATE]} changes accuracy negligibly across both models (Appendix~\ref{app:token_interface_controls}), separating token insertion from the subsequent activation-substitution effect.}

\paragraph{Position sensitivity under misaligned substitution}
We further examine whether {cross-framing substitution} depends on the structural placement of \texttt{[STATE]}. More than 95\% of generations under the tested misaligned placements either become incoherent or fail to produce a valid task-relevant response; the evaluated configurations, failure criterion, and canonical-position comparison are provided in Appendix~\ref{app:sub_position}.

\subsection{Dual Framing as a General Intervention Contrast}
\label{sec:dual_framing_steering}

{The preceding experiments characterize StateSwap within the dual-framing design. We next evaluate the \textsc{Sup}--\textsc{Elim} contrast independently of the introduced interface by removing \texttt{[STATE]} and instantiating the same contrast in a population-level CAA-style mean-difference steering framework. This analysis compares the direction derived from matched \textsc{Sup}--\textsc{Elim} prompts with the original CAA direction under otherwise matched settings.}

\begin{figure}[t]
\centering
\includegraphics[width=\columnwidth]{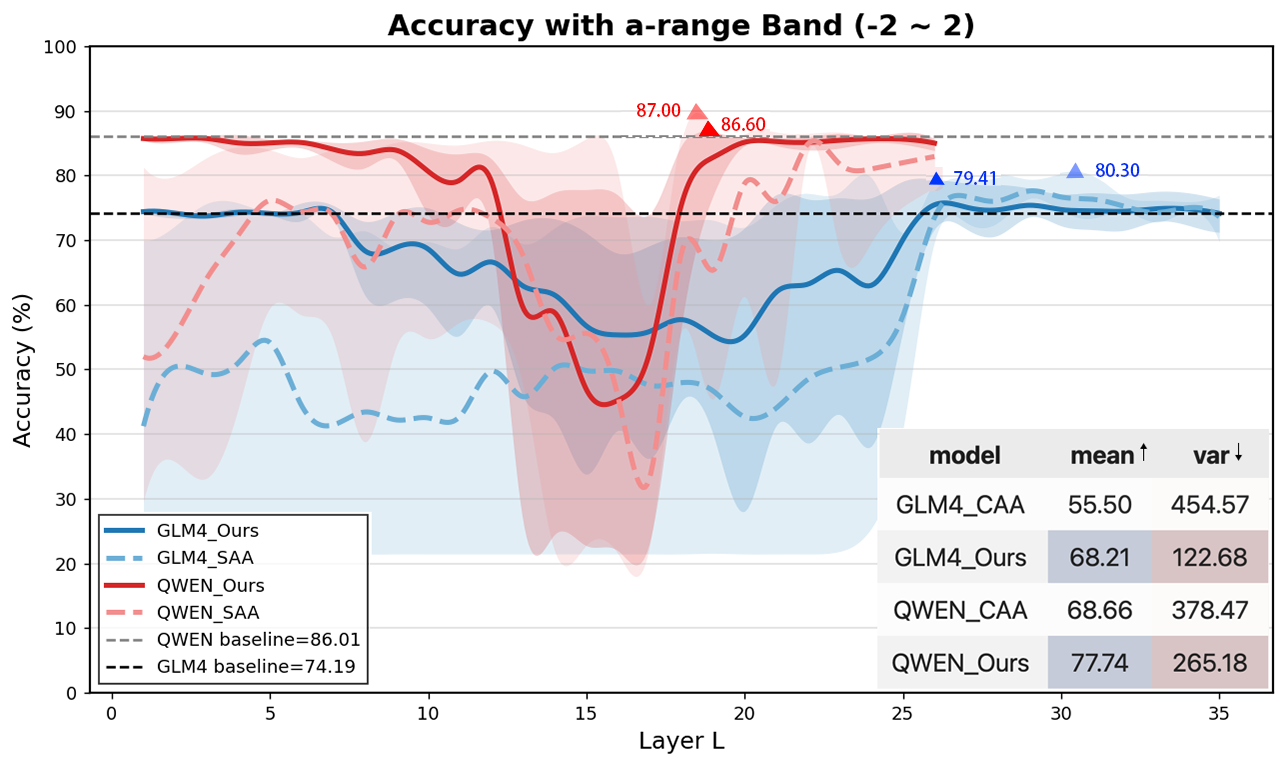}
\caption{
Accuracy under activation steering across intermediate layers.
Solid lines denote steering directions derived from the dual-framing contrast between \textsc{Sup} and \textsc{Elim} prompts, while dashed lines denote the original CAA baseline.
Red and blue curves correspond to Qwen-2.5-7B and GLM-4-9B, respectively.
Horizontal dashed lines indicate unsteered baseline accuracies.
}
\label{fig:compare_CAA}
\end{figure}

{As shown in Figure~\ref{fig:compare_CAA}, the dual-framing direction produces systematic layer-dependent effects in both models even without the \texttt{[STATE]} token. Detailed aggregate statistics and complete layer-wise results, including comparisons with the original CAA direction and failures under strong coefficients, are provided in Appendix~\ref{app:steering_results}.}

\section{Discussion}

The choice of \textsc{Sup} and \textsc{Elim} is central to what can be inferred from StateSwap. In preliminary analyses, we considered other complementary prompt pairs, including strictly translated Chinese and English instructions and prompts requesting a direct answer versus explicit reasoning. Such diversity can benefit prediction aggregating alternative reasoning paths improves accuracy \citep{wang2022self,lin2024just}. Language contrasts remained strongly classifiable from shallow through deep layers, plausibly because they introduce pervasive lexical and syntactic differences. Direct-answer and reasoning-first prompts were most distinguishable in intermediate and late layers, potentially reflecting different response strategies and generation plans. In both cases, a classifier could exploit factors broader than the decision distinction of interest.

The \textsc{Sup}--\textsc{Elim} contrast was selected to reduce these confounds. The paired prompts preserve the question, options, language, decoding procedure, and answer key while changing whether the model identifies the correct option or eliminates the incorrect ones. Although this does not guarantee that framing is the only information encoded at \texttt{[STATE]}, it supports a more specific conclusion: logically equivalent decision objectives can induce distinguishable intermediate representations whose exchange affects downstream predictions. 

Additional, \textsc{Sup} and \textsc{Elim} are broad observational labels that may each contain finer-grained states associated with domain, uncertainty, response strategy, or required deliberation. A model facing a medical question, for example, may adopt a more cautious state, potentially expressed through longer chains of thought, while remaining within the same \textsc{Sup} or \textsc{Elim} class. Our analysis marginalizes over such within-class structure because it asks whether one controlled framing distinction is detectable and behaviorally relevant. Hierarchical or multi-factor analyses could instead separate framing from domain, confidence, and reasoning strategy, and test which finer-grained states remain independently substitutable.

\section{Conclusion}
StateSwap provides a training-free framework for analyzing framing-sensitive representations through a dedicated residual-stream interface. Across two evaluated LLMs, support- and elimination-oriented prompts induce separable intermediate-layer \texttt{[STATE]} activations. Exchanging these activations systematically changes downstream predictions, while the controls distinguish structured substitution from token insertion, final-content-token replacement, unstructured perturbations, and prompt ensembling. These findings establish the behavioral relevance of framing-conditioned activations within the tested models, multiple-choice framing protocol, and deterministic decoding setting.

\section*{Limitations}

We conduct the experiments in a controlled setting with four-option multiple-choice questions and deterministic greedy decoding. This design helps isolate activation-level effects from sampling variance and makes BASE-to-SUB differences easier to attribute to the intervention. However, it also limits the scope of our claims. We do not evaluate whether the same framing-sensitive states persist under sampling-based decoding, longer multi-step generation, or open-ended tasks.

The current formulation is also tied to the structure of multiple-choice questions. In particular, \textsc{Sup} and \textsc{Elim} prompts define complementary option sets, and our evaluation relies on parsing outputs into correct and incorrect options. Extending StateSwap beyond MCQs would require paired prompt formulations with a shared evaluation target, as well as task-specific agreement metrics for open-ended outputs.

Our method further relies on an untrained \texttt{[STATE]} token inserted at a fixed position. Although different random initializations of this token produce negligible variation in our experiments, the inserted token may still introduce subtle distributional or representational shifts. We therefore do not claim that the observed effects are independent of the interface design or intervention site.

StateSwap also assumes white-box access to model activations and parameters, leaving its applicability to black-box models untested. Majority voting over paraphrased prompts is used only to reduce prompt-realization noise during evaluation, and is not part of the intervention method itself.

\section*{Ethical Statements}

This work does not involve human subjects, user studies, or the collection of personal data. All experiments are conducted offline on publicly available multiple-choice benchmarks, including MMLU and MedQA-CH, using openly released language models. The generated responses are synthetic benchmark outputs and are not associated with real individuals.

StateSwap is proposed as a diagnostic method for studying framing-sensitive internal representations under controlled inference settings. Although the intervention changes intermediate activations, our experiments are limited to semantically equivalent multiple-choice prompts and deterministic decoding. We do not present the method as a deployed alignment or decision-support system.

A potential risk of activation-level interventions is that similar techniques could be adapted for broader model steering, including biased or undesired behavioral changes. We therefore restrict our claims to offline analysis and report ablations to distinguish structured state substitution from generic perturbations. Since one benchmark is medical, we emphasize that the results should not be interpreted as evidence of clinical reliability or suitability for real-world medical decision making. Any use in high-stakes settings would require separate domain validation and safety evaluation.

We used AI-assisted tools responsibly to support language editing and manuscript preparation; all research ideas, analyses, results, and conclusions were developed and verified by the authors, who take full responsibility for the content of this paper.

\bibliography{custom}

\appendix
\numberwithin{table}{section}
\numberwithin{figure}{section}
\section{Methodological Details}

\subsection{Notation Summary}
\label{app:tb}
For clarity, we summarize the StateSwap-specific notation below. Standard
Transformer notation follows Section~\ref{sec:method_interface}; arguments
$(I,q)$ are omitted when they are clear from context.
\begin{center}
\small
\setlength{\tabcolsep}{4pt}
\begin{tabularx}{\columnwidth}{@{}lX@{}}
\toprule
Term / Symbol & Object (this paper) \\
\midrule
framing and instance & $I\in\{\textsc{Sup},\textsc{Elim}\}$; $\bar I$ is the complementary framing and $q$ is the shared question \\
framing-conditioned input & $\mathbf{x}_I(q)=\mathrm{Concat}(I,q,\texttt{[STATE]})$ \\
interface position & $t_S\triangleq T$, the final position in $\mathbf{x}_I(q)$ \\
interface state & $\mathbf{s}^{(l)}(I,q)=\mathbf{h}^{(l)}_{t_S}(I,q)\in\mathbb{R}^d$ \\
intervention region & contiguous $W\subseteq\{1,\dots,L\}$ \\
feature-window indices & length $w$, start $j\in\{0,\dots,d-w\}$, and evaluated size set $\mathcal{W}\subseteq\{1,\dots,d\}$ \\
feature-window slice & $\mathbf{s}^{(l)}_{[j:j+w)}\in\mathbb{R}^w$ \\
window-level diagnostic & $d_l(j,w)$, the paired Cohen's $d$ after scalarization; $d_l^{\mathrm{rand}}$ and $d_l^{\mathrm{md}}$ denote the two diagnostics \\
per-size layer intensity & $\tilde d_l(w)$, the mean of the $K$ largest values in $\{|d_l(j,w)|\}_{j=0}^{d-w}$ \\
aggregated layer intensity & $\tilde d_l=|\mathcal{W}|^{-1}\sum_{w\in\mathcal{W}}\tilde d_l(w)$ \\
state substitution & $\mathbf{s}^{(k)}(I,q)\leftarrow\mathbf{s}^{(k)}_{\mathrm{cache}}(\bar I,q)$ for $k\in W$ \\
\bottomrule
\end{tabularx}
\end{center}

\subsection{Construction and Alignment of the \texttt{[STATE]} Interface}
\label{app:state_token}

We introduce \texttt{[STATE]} as an additional special token in the tokenizer vocabulary. Concretely, we add \texttt{[STATE]} via the tokenizer interface: \texttt{tokenizer.add\_special\_tokens(·)}. 
After extending the tokenizer vocabulary, we resize the model's input embedding matrix to match the updated vocabulary size. All experiments are inference-only: model parameters are not updated, and the embedding vector corresponding to \texttt{[STATE]} remains at its initialization value throughout.

\paragraph{Token-Level Alignment}
\label{app:state_align}

Our methodology requires \texttt{[STATE]} to occupy the same absolute token index $t_S$ across $\textsc{Elim}$ and $\textsc{Sup}$ prompts for the same question $q$. We ensure this by token-level alignment of the instruction segment.
Let the full prompt be
\[
\mathbf{x}_I(q) = \mathrm{Concat}(I, Q, \texttt{[STATE]}).
\]
We construct the instruction segment so that it has the same token length for all instructions under the model tokenizer. For practical guidance, please refer to Appendix~\ref{sec:prompt_state_appendix}.

\begin{algorithm}[ht]
\caption{{Tokenize and Align Instructions}}
\label{alg:tokenize_align}
{
\begin{algorithmic}[1]

\Require
    Instructions list $\mathcal{I} = \{I_1, I_2, \dots, I_n\}$
\Ensure
    Aligned token lists $\mathcal{T}_{\mathrm{aligned}}$,
    padding sizes $\mathcal{P}$,
    target instruction length $L_{\mathrm{target}}$

\State Initialize empty lists $\mathcal{T}_{\mathrm{raw}}$ and $L_{\mathrm{base}}$

\For{each instruction $I$ in $\mathcal{I}$}
    \State $tokens \leftarrow \text{Tokenize}(I)$
    \State Append $tokens$ to $\mathcal{T}_{\mathrm{raw}}$
    \State Append $|tokens|$ to $L_{\mathrm{base}}$
\EndFor

\State $L_{\mathrm{target}} \leftarrow \max(L_{\mathrm{base}})$
\State Initialize empty lists $\mathcal{T}_{\mathrm{aligned}}$ and $\mathcal{P}$

\For{$i \leftarrow 1$ to $n$}
    \State $pad_i \leftarrow L_{\mathrm{target}} - L_{\mathrm{base}}[i]$
    \State $aligned_i \leftarrow \mathrm{Concat}([\texttt{PAD}]^{pad_i},\mathcal{T}_{\mathrm{raw}}[i])$
    \State Append $aligned_i$ to $\mathcal{T}_{\mathrm{aligned}}$
    \State Append $pad_i$ to $\mathcal{P}$
\EndFor

\State \Return $\mathcal{T}_{\mathrm{aligned}}, \mathcal{P}, L_{\mathrm{target}}$

\end{algorithmic}
}
\end{algorithm}

Because the question block and the final \texttt{[STATE]} token are identical {in both framings}, equalizing the instruction-segment length guarantees that the absolute index $t_S$ of \texttt{[STATE]} is matched {between} $\textsc{Elim}$ and $\textsc{Sup}$ for each paired question $q$.

\subsection{Window Diagnostics and Candidate Region Selection}
\label{app:cohen_d}

For window size $w$, we use start indices $j \in \{0,\dots,d-w\}$ and the slice notation $[j:j+w)$.
For a scalarization $\phi:\mathbb{R}^w\rightarrow\mathbb{R}$, define paired differences for each question $i$:
\[
\Delta^{(l,j,w)}_i
= \phi(\mathbf{s}^{(l)}_{\textsc{Elim},i,[j:j+w)})
 - \phi(\mathbf{s}^{(l)}_{\textsc{Sup},i,[j:j+w)}).
\]
We report the paired Cohen's d statistic:
\begin{equation}
\label{eq:cohen_d}
d_l(j,w)=\frac{\overline{\Delta^{(l,j,w)}}}{\mathrm{std}(\Delta^{(l,j,w)})+\epsilon},
\end{equation}
where the mean and standard deviation are computed over $\{\Delta^{(l,j,w)}_i\}_{i=1}^N$.

\paragraph{Candidate Region Selection}
\label{app:region_select}

We compute layer-level intensities from absolute window-level separability scores using a
Top-$K$ mean aggregation.
Because both diagnostics are reported in experiments, the same procedure is applied
independently to $d^{\mathrm{rand}}_l(j,w)$ and $d^{\mathrm{md}}_l(j,w)$.

For a given diagnostic $d_l(j,w)$ and window size $w$, we define
\[
\tilde d_l(w)=\mathrm{TopKMean}\Big(\{|d_l(j,w)|\}_{j=0}^{d-w};\,K\Big),
\]
where $\mathrm{TopKMean}(\cdot;K)$ averages the $K$ largest values.
Unless stated otherwise, we further average over a small set of window sizes $\mathcal{W}$:
\[
\tilde d_l=\frac{1}{|\mathcal{W}|}\sum_{w\in\mathcal{W}}\tilde d_l(w).
\]

Layers are then ranked by $\tilde d_l$.
To determine a threshold for selecting candidate layers, we rely on an empirical consistency
criterion guided by an independent PCA analysis.
Specifically, we observe that layers whose $\tilde d_l$ values exceed approximately
$80\%$ of the maximum separability score consistently coincide with the region showing
the clearest separation in the PCA visualization.
Based on this observation, we select layers satisfying
\[
\tilde d_l \geq 0.8 \cdot \max_{l'} \tilde d_{l'}
\]
For each evaluated model, the layers satisfying this criterion form a single contiguous interval, which we denote by $W$. This threshold is not tuned for downstream performance. It is fixed across models and datasets and serves only as a diagnostic criterion for localizing layers that exhibit stable, representation-level differences {between the two framings}.

\section{Experimental and Evaluation Details}

\subsection{Datasets and Data Splits}
\label{app:data_stats}

We evaluate our approach on two four-option multiple-choice benchmarks: MedQA-CH and a reasoning-focused subset of MMLU. All downstream inference and reported task evaluations are conducted on the official test splits. A separate set of 50 examples from the training split is used only to localize the candidate intervention window; these examples do not enter the reported test evaluation. Appendix~\ref{app:window_robustness} provides the full calibration protocol and a larger-scale bootstrap robustness analysis.

\paragraph{MedQA-CH}
MedQA-CH is the Chinese subset of MedQA, a medical question answering benchmark designed to evaluate professional-level clinical reasoning. It consists of 1,015 multiple-choice questions covering clinical knowledge and diagnostic reasoning. All questions are presented in Chinese and evaluated independently from English benchmarks.

\paragraph{MMLU (Reasoning-Focused Subset)}
We construct a reasoning-oriented subset of MMLU covering 17 subject categories that emphasize logical reasoning, mathematical thinking, legal analysis, ethical judgment, and domain-specific inference. The selected subsets span formal logic, mathematics, law, philosophy, medicine, and social sciences. All MMLU evaluations are conducted in English using the official test splits.

Table~\ref{tab:dataset_stats} summarizes the detailed dataset composition. In total, our evaluation comprises 5,622 multiple-choice questions. For bilingual evaluation, English and Chinese benchmarks are tested separately rather than mixed within a single evaluation setting.

\begin{table}[htbp]
\centering
\small
\begin{tabular}{l r}
\hline
\textbf{Dataset / Subset} & \textbf{\# Questions} \\
\hline
\multicolumn{2}{l}{\textit{MedQA}} \\
\hline
MedQA-CH & 1,015 \\
\hline
\multicolumn{2}{l}{\textit{MMLU (Reasoning-Focused Subset)}} \\
\hline
Formal Logic & 125 \\
Logical Fallacies & 163 \\
College Math Logic & 100 \\
Abstract Algebra (Proofs) & 100 \\
High School Math & 270 \\
Professional Law & 1,534 \\
Jurisprudence & 108 \\
Moral Disputes & 346 \\
Business Ethics & 100 \\
Philosophy & 311 \\
Sociology & 201 \\
World History & 237 \\
U.S. History & 204 \\
World Religions & 171 \\
Bioethics & 272 \\
Clinical Knowledge (Reasoning) & 265 \\
Medical Genetics & 100 \\
\hline
Total (MMLU subset) & 4,607 \\
\hline
\textbf{Total (All datasets)} & \textbf{5,622} \\
\hline
\end{tabular}
\caption{Dataset statistics. All datasets are evaluated using their official test splits. MedQA-CH is evaluated in Chinese, while MMLU subsets are evaluated in English.}
\label{tab:dataset_stats}
\end{table}

\subsection{Decoding and Inference Configuration}
\label{app:hyperparameters}
All experiments are conducted in a zero-shot setting using deterministic greedy decoding. Specifically, we set \texttt{do\_sample=False} and \texttt{max\_new\_tokens=1024}. As greedy decoding is used, sampling-related parameters such as \texttt{temperature} and \texttt{top\_k} are not applicable and are ignored by the generation backend. All random seed values are 123.

The maximum context length is fixed across all experiments, and inputs exceeding this limit are truncated accordingly. Right padding is applied with a maximum of 64 padding tokens. All generations use the official chat templates provided by the respective models without modification.

For reproducibility, all runs are executed with fixed random seeds and deterministic computation settings. Inference is performed using \texttt{torch\_dtype=float16}, with key–value caching enabled to ensure consistent decoding behavior across runs.

\subsection{Detailed Prompt Templates}
\label{sec:prompt_state_appendix}

This section provides the prompt templates used in the experiments.

\paragraph{Prompt Components}
Each input prompt consists of three components: an instruction segment $I$, a question segment $Q$ that contains the question text, and an option segment $O$ that enumerates all candidate answers in a fixed format. In addition, a special interface token \texttt{[STATE]} is appended to the end of the prompt to mark the position at which the model is expected to produce its output. For models that employ a beginning-of-sequence token, the tokenizer’s default behavior is preserved.

\paragraph{Instruction Templates}
{We use two groups of framing-specific instruction templates with aligned task objectives.} One group instructs the model to reason step by step and identify \textbf{correct} options (Group 1), while the other instructs the model to reason step by step and identify \textbf{incorrect} options (Group 2). Within each group, multiple paraphrased variants are used to reduce dependence on a specific wording. All instruction templates are fixed across datasets and models.

\begin{boxH}
\textbf{Group 1 - Sup prompt}

Instruction: Proceed step by step: first list the complete chain of reasoning, then pick out \textbf{Correct} options. Question: \texttt{[QUESTION]}

Instruction: Take a careful, step-by-step approach. First, reason in detail, then find \textbf{Correct} choices. Question: \texttt{[QUESTION]}
\end{boxH}

\begin{boxH}
\textbf{Group 2 - Elim prompt}

Instruction: Proceed step by step: first list the complete chain of reasoning, then pick out \textbf{Wrong} options. Question: \texttt{[QUESTION]}

Instruction: Take a careful, step-by-step approach. First, reason in detail, then find \textbf{Incorrect} choices. Question: \texttt{[QUESTION]}
\end{boxH}

For {each framing}, we instantiate two instruction templates, yielding two deterministic greedy predictions for \textsc{Sup} and two for \textsc{Elim}. Unless otherwise specified, reported ensemble results aggregate these paired prompt realizations by strict-majority voting within the corresponding {framing} or StateSwap variant; a 1--1 tie is counted as an incorrect ensemble prediction. This protocol separates prompt-template variation from decoding randomness, since all runs use greedy decoding.

\subsection{Robustness to \texttt{[STATE]} Initialization}
\label{app:state_init_robustness}

The purpose of this experiment is to verify that StateSwap does not rely on incidental randomness from the untrained \texttt{[STATE]} token.
Table~\ref{tab:state_init_full} shows that output similarity remains near-perfect across different random initializations, supporting the claim that \texttt{[STATE]} functions as a deterministic state interface rather than a source of stochastic variation.

\begin{table*}[t]
\centering
\small
\begin{tabular}{lrrcccc}
\toprule
\multirow{2}{*}{Swap direction} & \multirow{2}{*}{Seed$_a$} & \multirow{2}{*}{Seed$_b$} &
\multicolumn{2}{c}{LA-cos $\uparrow$} & \multicolumn{2}{c}{BERT-F1 $\uparrow$} \\
\cmidrule(lr){4-5}\cmidrule(lr){6-7}
 &  &  & mean & std & mean & std \\
\midrule
\multirow{10}{*}{\makecell[c]{$Elim\_Baseline$\\$\downarrow$ \\$Sup\_Baseline$}}
 & 1 & 2   & 9.99999986589e-01 & 7.81e-08 & 9.99999991059e-01 & 5.65e-08 \\
 & 1 & 3   & 9.99999986589e-01 & 7.81e-08 & 9.99999991059e-01 & 5.65e-08 \\
 & 1 & 12  & 9.99999989569e-01 & 7.38e-08 & 9.99999991059e-01 & 5.65e-08 \\
 & 1 & 123 & 9.99999989569e-01 & 7.38e-08 & 9.99999991059e-01 & 5.65e-08 \\
 & 2 & 3   & 9.99999986589e-01 & 7.81e-08 & 9.99999991059e-01 & 5.65e-08 \\
 & 2 & 12  & 9.99999989569e-01 & 7.38e-08 & 9.99999991059e-01 & 5.65e-08 \\
 & 2 & 123 & 9.99999989569e-01 & 7.38e-08 & 9.99999991059e-01 & 5.65e-08 \\
 & 3 & 12  & 9.99999989569e-01 & 7.38e-08 & 9.99999991059e-01 & 5.65e-08 \\
 & 3 & 123 & 9.99999989569e-01 & 7.38e-08 & 9.99999991059e-01 & 5.65e-08 \\
 & 12 & 123& 9.99999991059e-01 & 7.34e-08 & 9.99999991059e-01 & 5.65e-08 \\
\midrule
\multirow{10}{*}{\makecell[c]{$Sup\_Baseline$\\$\downarrow$ \\$Elim\_Baseline$}}
 & 1 & 2   & 9.99999982119e-01 & 7.04e-08 & 9.99999976158e-01 & 8.19e-08 \\
 & 1 & 3   & 9.99999982119e-01 & 7.04e-08 & 9.99999976158e-01 & 8.19e-08 \\
 & 1 & 12  & 9.99999984354e-01 & 6.62e-08 & 9.99999976158e-01 & 8.19e-08 \\
 & 1 & 123 & 9.99999984354e-01 & 6.62e-08 & 9.99999976158e-01 & 8.19e-08 \\
 & 2 & 3   & 9.99999982119e-01 & 7.04e-08 & 9.99999976158e-01 & 8.19e-08 \\
 & 2 & 12  & 9.99999984354e-01 & 6.62e-08 & 9.99999976158e-01 & 8.19e-08 \\
 & 2 & 123 & 9.99999984354e-01 & 6.62e-08 & 9.99999976158e-01 & 8.19e-08 \\
 & 3 & 12  & 9.99999984354e-01 & 6.62e-08 & 9.99999976158e-01 & 8.19e-08 \\
 & 3 & 123 & 9.99999984354e-01 & 6.62e-08 & 9.99999976158e-01 & 8.19e-08 \\
 & 12 & 123& 9.99999985844e-01 & 6.58e-08 & 9.99999976158e-01 & 8.19e-08 \\
\bottomrule
\end{tabular}
\caption{Detailed robustness results for random \texttt{[STATE]} initialization.
We sample 40 instances and rerun inference with five random seeds (1, 2, 3, 12, 123), keeping all other settings fixed.
For each swap direction and seed pair, we report the mean and standard deviation of LA-cos and BERT-F1 computed over the 40 instances, showing negligible variation across seeds.}
\label{tab:state_init_full}
\end{table*}

\subsection{Output Parsing and Evaluation Metrics}
\label{app:metrics_parsing}

\subsubsection{Parsing to Option Sets}
\label{app:parsing}

To ensure consistency across languages and facilitate automatic evaluation, model outputs are post-processed into a standardized representation consisting of two fields: \texttt{Correct options:[\dots]} and \texttt{Wrong options:{[\dots]}}. This step preserves the semantic meaning of the original responses while enforcing a fixed, machine-readable output format.

\paragraph{Options Answer extraction}
Let $\mathcal{O}=\{A,B,C,D\}$ denote the option alphabet.
We parse each field into a set $S \subseteq \mathcal{O}$ by extracting all uppercase letters in
$\mathcal{O}$ from the field, ignoring punctuation and separators (e.g., commas, Chinese enumeration marks).

\subsubsection{Metric Definitions}
\label{app:metrics}

\paragraph{Accuracy (ACC)}
For each question with ground-truth correct option $y \in \mathcal{O}$, the prediction under a given projection is considered correct if $y \in S$, where $S$ denotes the set of options predicted as correct by the model. Formally, accuracy is defined as:
\[
\mathrm{ACC} = \frac{1}{N} \sum_{i=1}^{N} \mathbb{I}\left[y_i \in S_i\right],
\]
where $N$ is the total number of questions, $y_i$ is the ground-truth correct option for the $i$-th question, $S_i$ denotes the predicted set of correct options, and $\mathbb{I}[\cdot]$ is the indicator function.

\paragraph{Jaccard}
\label{app:jaccard}
Let $C_{\text{Sup}}$ and $C_{\text{Elim}}$ denote the sets of question indices for which the model predicts the correct answer under the {support- and elimination-oriented framings}, respectively. For each question, correctness is determined solely based on the \texttt{Correct options} field in the normalized output, as described above.

We define {cross-framing} agreement using the Jaccard index over these two sets:
\[
J = \frac{|C_{\text{Sup}} \cap C_{\text{Elim}}|}{|C_{\text{Sup}} \cup C_{\text{Elim}}|}.
\]

Intuitively, a question contributes to the intersection if and only if the correct option appears in the \texttt{Correct options} list under {both framings}. For example, if the \textsc{Sup} output is \texttt{Correct: [1,0,0,0]} and the \textsc{Elim} output is also \texttt{Correct: [1,0,0,0]}, the question is counted as overlapping; whereas if the two outputs differ (e.g., \texttt{[1,0,0,0]} vs.\ \texttt{[0,1,0,0]}), it is not. This metric therefore measures agreement in successful decisions {between the two framings}, rather than overlap between option sets for individual questions.

\paragraph{Relation Between ACC and Jaccard}
{While ACC measures whether the model produces the correct answer under a single framing, the Jaccard index captures the consistency of correct decisions between the two framings. In particular, ACC reflects per-framing performance, whereas Jaccard evaluates cross-framing agreement conditioned on correctness. A high ACC with a low Jaccard score indicates that the model can answer correctly under either framing but does so inconsistently, while a high Jaccard score reflects stable decision-making between the two framings.} The two metrics therefore provide complementary views of model behavior.

\paragraph{Length-Aware Cosine Similarity}
Given a baseline response $r$ and its perturbed counterpart $\tilde{r}$, we compute the cosine similarity between their sentence embeddings. To account for potential length mismatches, we apply a length-aware scaling factor:
\[
\mathrm{LAC}(r, \tilde{r}) = \cos(\mathbf{e}_r, \mathbf{e}_{\tilde{r}}) \cdot \frac{\min(|r|, |\tilde{r}|)}{\max(|r|, |\tilde{r}|)},
\]
where $\mathbf{e}_r$ and $\mathbf{e}_{\tilde{r}}$ denote normalized sentence embeddings, and $|r|$ denotes the character length of the response. 

This metric captures semantic similarity while penalizing excessive length divergence caused by perturbations. The length-aware scaling term is introduced to account for cases where perturbations to the hidden state cause the model to generate excessively long or verbose outputs that are only weakly related to the original response. In such cases, standard cosine similarity alone may overestimate semantic consistency due to shared topical content. By down-weighting similarity scores when large length discrepancies occur, this metric penalizes uncontrolled or drifting generations and better reflects whether the core semantic content is preserved.

\paragraph{BERTScore}
We additionally compute BERTScore between the perturbed and baseline outputs, reporting F1 scores. BERTScore measures token-level semantic similarity using contextualized embeddings and is sensitive to both content preservation and phrasing changes. In our experiments, the baseline output is treated as the reference and the perturbed output as the candidate.

\paragraph{Relation Between LAC and BERTScore}
These two metrics are computed by comparing each perturbed output against its corresponding baseline output without any state modification. They are therefore intended to measure the sensitivity of the generation to perturbations applied to the \texttt{[STATE]} representation, rather than absolute output quality.

Length-aware cosine similarity reflects global semantic consistency between the perturbed and baseline responses, while BERTScore captures fine-grained lexical and contextual deviations. Importantly, higher scores do not necessarily indicate better performance: values close to 1 suggest that the perturbation has little effect on the output, whereas lower scores indicate that modifying the internal state leads to substantial changes in the generated content. Extremely low scores may further reflect degenerate or incoherent outputs caused by the perturbation.

\subsubsection{Error-Elimination Metric}
\label{app:error_metric}

For instance $i$ and {framing} $p\in\{\textsc{Sup},\textsc{Elim}\}$, let $W_i^{(p)}\subseteq\mathcal O$ be the options parsed from the \texttt{Wrong options} field. We define the eliminated-incorrect count as
\[
E_i^{(p)}=\bigl|W_i^{(p)}\setminus\{y_i\}\bigr|\in\{0,1,2,3\}.
\]

Let \(E_{i,\text{base}}^{(p)}\) and \(E_{i,\text{sub}}^{(p)}\) denote the values before and after substitution. We define
\[
\Delta_i^{(p)} = E_{i,\text{sub}}^{(p)} - E_{i,\text{base}}^{(p)}.
\]
An instance is counted as a \textsc{Right Shift} if \(\Delta_i^{(p)}>0\), and as a \textsc{Left Shift} if \(\Delta_i^{(p)}<0\). We report the proportions of right- and left-shifted instances under {each framing} and their macro-average across \textsc{Sup} and \textsc{Elim}. A higher right-shift frequency indicates stronger incorrect-option elimination.

\section{Additional Behavioral Results}
\subsection{Ensemble and Per-Task Results}
\label{app:ensemble_ablation}

Our main reported results are based on deterministic ensemble inference rather than a single prompt realization. Specifically, each {single-framing} result is averaged over two greedy runs with different prompt realizations. To further examine whether the observed gains can be explained by simple aggregation effects, we compare several ensemble variants: {single-framing} baselines (\textsc{Sup}-only and \textsc{Elim}-only), StateSwap variants (\textsc{Sup}(\textsc{Elim} State) and \textsc{Elim}(\textsc{Sup} State)), a naive {cross-framing} ensemble (\textsc{Sup}+\textsc{Elim}), {framing-plus-StateSwap} ensembles, and a dual StateSwap ensemble. As shown in Table~\ref{tab:ensemble_ablation}, ensembles involving swapped \texttt{[STATE]} predictions consistently outperform ensembles formed only from the original {framings}. In particular, the dual StateSwap ensemble achieves the best accuracy for both Qwen-2.5-7B and GLM-4-9B, suggesting that StateSwap transfers useful decision-state information beyond ordinary variance reduction from ensembling.

\begin{table}[t]
\centering
\small
\begin{tabular}{lcc}
\toprule
\textbf{Method} & \makecell{\textbf{Qwen-2.5-7B}\\\textbf{Acc.}} & \makecell{\textbf{GLM-4-9B}\\\textbf{Acc.}} \\
\midrule

\multicolumn{3}{c}{\textit{2-Run Ensemble}} \\
\midrule

\textsc{Elim}-only 
& 0.7436 & 0.7354 \\

\textsc{Elim}(\textsc{Sup} State)
& \textbf{0.7745} & \textbf{0.7792} \\
\midrule
\textsc{Sup}-only
& 0.7712 & 0.7591 \\

\textsc{Sup}(\textsc{Elim} State)
& \textbf{0.8026} & \textbf{0.7682} \\

\midrule

\multicolumn{3}{c}{\textit{4-Run Ensemble}} \\
\midrule

\makecell[l]{
\textsc{Sup} + \\
\textsc{Sup}(\textsc{Elim} State)
}
& 0.7939 & 0.7692 \\

\makecell[l]{
\textsc{Elim} + \\
\textsc{Elim}(\textsc{Sup} State)
}
& 0.7844 & 0.7861 \\

\midrule
\textsc{Sup} + \textsc{Elim} 
& 0.8020 & 0.7890 \\

\makecell[l]{
\textsc{Sup}(\textsc{Elim} State) + \\
\textsc{Elim}(\textsc{Sup} State)
}
& \textbf{0.8156} & \textbf{0.7949} \\

\bottomrule
\end{tabular}

\caption{
Ensemble ablation results. The 2-run setting reports deterministic greedy ensembles within a {single framing} or StateSwap condition. The 4-run setting combines two 2-run groups. StateSwap-based ensembles outperform the naive \textsc{Sup}+\textsc{Elim} ensemble on both models.
}
\label{tab:ensemble_ablation}
\end{table}

Since combining {two framings} can produce an even number of votes, we further evaluate a weighted majority rule for tie-breaking. We use a linear vote of the form $\mathrm{vote}=\textsc{Sup}+\alpha\cdot\textsc{Elim}$ and search over $\alpha$ for 30 trials, reporting the best achievable accuracy for each ensemble. As shown in Table~\ref{tab:weighted_vote_ablation}, even under this optimized aggregation scheme, the dual StateSwap ensemble remains stronger than direct \textsc{Sup}--\textsc{Elim} ensembling. This indicates that the StateSwap gains cannot be explained away by a better voting heuristic alone.

\begin{table}[t]
\centering
\small
\begin{tabular}{lcc}
\toprule
\textbf{Method} & \makecell{\textbf{Qwen-2.5-7B}\\\textbf{Acc.}} & \makecell{\textbf{GLM-4-9B}\\\textbf{Acc.}} \\
\midrule

\multicolumn{3}{c}{\textit{4-Run Ensemble with Weighted Voting}} \\
\midrule

\makecell[l]{
\textsc{Sup} + \\
\textsc{Sup}(\textsc{Elim} State)
}
& 0.8008 & 0.7732 \\
\midrule
\makecell[l]{
\textsc{Elim} + \\
\textsc{Elim}(\textsc{Sup} State)
}
& 0.7943 & 0.7861 \\
\midrule
\textsc{Sup} + \textsc{Elim}
& 0.8059 & 0.7939 \\
\midrule
\makecell[l]{
\textsc{Sup}(\textsc{Elim} State) + \\
\textsc{Elim}(\textsc{Sup} State)
}
& \textbf{0.8225} & \textbf{0.8018} \\

\bottomrule
\end{tabular}

\caption{
Weighted-vote ensemble ablation. We search over the linear tie-breaking weight $\alpha$ for 30 trials and report the best accuracy. StateSwap-based ensembling remains stronger than direct \textsc{Sup}+\textsc{Elim} aggregation.
}
\label{tab:weighted_vote_ablation}
\end{table}

\paragraph{Per-task MMLU accuracy} Table~\ref{tab:mmlu_swap_reordered} reports per-task accuracy under {the \textsc{Sup} and \textsc{Elim} framings}.
Compared with the Base setting, the Sub variant achieves consistent or improved performance
across most MMLU categories.

\begin{table*}[t]
\centering
\small
\setlength{\tabcolsep}{4pt}
\begin{tabular}{l|cccc|cccc}
\hline
\multirow{2}{*}{Task} 
& \multicolumn{4}{c|}{\textbf{Qwen-2.5-7B}} 
& \multicolumn{4}{c}{\textbf{GLM-4-9B}} \\
\cline{2-9}
& Base-Elim & Sub-Elim & Base-Sup & Sub-Sup 
& Base-Elim & Sub-Elim & Base-Sup & Sub-Sup \\
\hline
abstract\_algebra\_proofs & 63.27 & 61.62 & 63.64 & 66.00 & 65.31 & 65.31 & 59.60 & 70.71 \\
bioethics                & 76.01 & 79.04 & 78.31 & 80.88 & 81.18 & 86.40 & 82.35 & 87.50 \\
business\_ethics         & 71.72 & 71.00 & 68.00 & 68.00 & 69.39 & 73.47 & 68.69 & 74.75 \\
clinical\_knowledge      & 75.19 & 74.05 & 69.96 & 69.81 & 74.90 & 78.03 & 73.86 & 76.23 \\
college\_math            & 70.97 & 69.47 & 61.05 & 63.00 & 63.04 & 72.34 & 58.95 & 61.46 \\
formal\_logic            & 64.46 & 63.64 & 62.81 & 63.93 & 61.29 & 68.25 & 64.80 & 65.08 \\
hs\_math                 & 86.92 & 88.67 & 75.00 & 77.61 & 81.42 & 83.90 & 62.20 & 70.45 \\
jurisprudence            & 71.30 & 71.96 & 69.44 & 68.52 & 80.37 & 79.63 & 75.93 & 75.00 \\
logical\_fallacies       & 81.51 & 80.13 & 81.87 & 84.28 & 77.07 & 85.19 & 83.23 & 83.44 \\
medical\_genetics        & 84.38 & 84.85 & 80.00 & 80.00 & 85.00 & 86.00 & 80.00 & 81.00 \\
moral\_disputes          & 61.70 & 62.39 & 63.85 & 65.51 & 69.01 & 72.09 & 63.58 & 65.90 \\
philosophy               & 64.61 & 65.48 & 71.70 & 73.31 & 68.93 & 71.06 & 63.67 & 69.45 \\
professional\_law        & 47.22 & 48.92 & 50.20 & 50.36 & 54.46 & 57.67 & 53.82 & 55.71 \\
sociology                & 73.06 & 71.94 & 71.28 & 71.43 & 76.02 & 84.26 & 72.73 & 75.00 \\
us\_history              & 85.05 & 85.20 & 80.20 & 82.14 & 84.92 & 89.11 & 85.93 & 87.13 \\
world\_history           & 81.62 & 79.24 & 84.55 & 82.63 & 84.85 & 86.38 & 86.38 & 86.44 \\
world\_religions         & 85.71 & 84.52 & 82.53 & 82.63 & 84.02 & 86.90 & 81.66 & 82.25 \\
\hline
\end{tabular}
\caption{
Extended per-task results on the MMLU benchmark.
We report accuracy under {the two framings} (Elim and Sup),
each evaluated with the original model (Base) and with \texttt{[STATE]} substitution (Sub).
This table complements the main results by providing fine-grained, task-level comparisons.
All values are percentages(\%); higher is better.
}
\label{tab:mmlu_swap_reordered}
\end{table*}

\subsection{Decision Determinacy Behavior}
\label{sec:decision_determinacy}

Beyond answer accuracy, we analyze whether \texttt{[STATE]} substitution alters the model’s determinacy in eliminating incorrect options, using the error-elimination metric defined in Appendix~\ref{app:error_metric}.

As shown in Table~\ref{tab:error_shift}, right shifts consistently occur more frequently than left shifts across models and datasets, indicating that \texttt{[STATE]} substitution more often strengthens the model’s ability to eliminate incorrect options, rather than relaxing it.
\begin{table}[t]
\centering
\small
\begin{tabular}{l l c c c}
\toprule
\textbf{Model} & \textbf{Dataset} & Right Shift &  & Left Shift \\
\midrule
\multirow{2}{*}{Qwen-2.5-7B}
 & MMLU  & 7.25\%  & $>$ & 7.17\% \\
 & MedQA & 11.31\% & $>$ & 9.29\% \\
\midrule
\multirow{2}{*}{GLM-4-9B}
 & MMLU  & 9.75\%  & $>$ & 7.52\% \\
 & MedQA & 15.24\% & $>$ & 11.39\% \\
\bottomrule
\end{tabular}
\caption{Directional changes in eliminated-incorrect mass from baseline to substitution (proportions over all instances).
Right Shift indicates an increase in eliminated-incorrect mass, while Left Shift indicates a decrease.}
\label{tab:error_shift}
\end{table}

This asymmetric shift aligns with the gains observed in ACC~Elim and Jaccard (Table~\ref{tab:acc_jaccard_main}), suggesting that the observed accuracy improvements are driven by enhanced error-elimination capability, manifested as more decisive and systematic suppression of incorrect options, rather than random variation or superficial answer changes.

\subsection{Detailed Steering Results}
\label{app:steering_results}

{
This experiment isolates how a population-level steering direction changes when the contrast is constructed from instructions rather than answer options. It contains no \texttt{[STATE]} token and does not perform instance-level state substitution.

\paragraph{Matched contrast construction}
For calibration example $i$, let $p_i^{\textsc{Sup}}$ request the correct option, let $p_i^{\textsc{Elim}}$ request an incorrect option, and let $y_i$ be the correct answer letter. We pair $y_i$ with a deterministic incorrect letter $\widetilde{y}_i$, chosen as the first letter in $\{A,B,C,D\}$ that differs from $y_i$. Let $h_l(p,y)$ denote the residual-stream activation at the appended answer token $y$ after layer $l$. Standard CAA holds the instruction fixed and changes the completion:
\begin{equation}
v_l^{\mathrm{CAA}}
=\frac{1}{N}\sum_{i=1}^{N}
\left[h_l(p_i^{\textsc{Sup}},y_i)-h_l(p_i^{\textsc{Sup}},\widetilde{y}_i)\right].
\label{eq:caa_option_direction}
\end{equation}
The dual-framing construction holds the completion fixed and changes the instruction:
\begin{equation}
v_l^{\mathrm{DF}}
=\frac{1}{N}\sum_{i=1}^{N}
\left[h_l(p_i^{\textsc{Sup}},y_i)-h_l(p_i^{\textsc{Elim}},y_i)\right].
\label{eq:dual_instruction_direction}
\end{equation}
We use $N=50$ calibration examples for both directions. The two contrasts share the positive pair $(p_i^{\textsc{Sup}},y_i)$. Their negative members are complementary counterfactuals: CAA preserves the instruction but forces an inconsistent option, whereas dual framing preserves the option but forces an inconsistent instruction--completion pair. Neither negative member is treated as a valid answer to its instruction. Both serve only to define a matched activation difference.

\paragraph{Steering and evaluation protocol}
We independently $\ell_2$-normalize each layer-wise direction, giving $\widehat{v}_l=v_l/\lVert v_l\rVert_2$. For one layer at a time, we modify the residual stream at the answer boundary as
\begin{equation}
h_l \leftarrow h_l + a\widehat{v}_l,
\qquad a\in\{-2,-1,0,1,2\}.
\label{eq:global_steering_intervention}
\end{equation}
Both methods use identical hooks, prompt templates at evaluation, coefficient values, and next-token scoring over $\{A,B,C,D\}$. We evaluate all 1,015 MedQA-CH test questions with Qwen-2.5-7B and GLM-4-9B. Accuracy at $a=0$ is the unsteered reference and is omitted from Table~\ref{tab:full_steering_results}. We sweep Layers 1--26 for Qwen and Layers 1--35 for GLM. Consequently, the comparison changes only whether the calibration contrast perturbs the answer option or the instruction.

\paragraph{Aggregate comparison}
Averaged across the evaluated layers, the dual-framing direction retains higher accuracy than CAA at $a=-2$, $-1$, and $2$. The respective dual-framing and CAA means are 59.03 vs.\ 37.16, 69.67 vs.\ 48.85, and 68.15 vs.\ 60.35 for GLM. They are 71.21 vs.\ 52.26, 81.61 vs.\ 75.97, and 74.77 vs.\ 62.90 for Qwen. At $a=1$, the means are nearly identical: 74.05 vs.\ 74.07 for GLM and 83.37 vs.\ 83.52 for Qwen.

The clearest stability difference occurs at $a=-1$. Relative to CAA, the layer-wise range narrows from 50.54 to 19.90 points for GLM and from 60.89 to 28.08 points for Qwen. At $a=1$, the range is similar for GLM (10.93 for CAA and 9.16 for dual framing) and narrower under dual framing for Qwen (27.00 vs.\ 12.41). Strong coefficients still cause sharp failures at individual layers. Higher retained accuracy also measures robustness to intervention, not greater steering strength by itself. The evidence therefore supports a more bounded layer-wise response under this matched protocol, rather than general superiority over CAA. Table~\ref{tab:full_steering_results} reports every layer--coefficient result.
}

\begin{table*}[p]
\centering
\small
\renewcommand{\arraystretch}{0.88}
\begin{tabular*}{\textwidth}{@{\extracolsep{\fill}}l|rrrr|rrrr@{}}
\toprule
\multirow{2}{*}{Layer}
& \multicolumn{4}{c|}{{\textbf{Two-framing}}}
& \multicolumn{4}{c}{\textbf{CAA}} \\
\cmidrule(lr){2-5}\cmidrule(lr){6-9}
& $a=-2$ & $a=-1$ & $a=1$ & $a=2$
& $a=-2$ & $a=-1$ & $a=1$ & $a=2$ \\
\midrule
\multicolumn{9}{c}{\textbf{GLM-4-9B}} \\
\midrule
1 & 74.68 & 74.19 & 74.38 & 74.38 & 21.48 & 46.60 & 69.95 & 26.90 \\
2 & 73.40 & 73.89 & 74.68 & 74.68 & 22.46 & 58.13 & 72.22 & 48.87 \\
3 & 73.40 & 73.50 & 74.29 & 74.29 & 22.17 & 50.15 & 75.67 & 56.33 \\
4 & 74.88 & 74.48 & 74.09 & 73.79 & 21.48 & 37.14 & 75.76 & 69.75 \\
5 & 74.19 & 74.19 & 74.19 & 73.89 & 23.94 & 50.44 & 74.38 & 67.98 \\
6 & 75.76 & 75.37 & 73.20 & 73.00 & 21.48 & 33.20 & 74.09 & 48.87 \\
7 & 74.48 & 75.37 & 73.69 & 73.30 & 21.48 & 35.67 & 70.54 & 38.03 \\
8 & 61.48 & 69.56 & 72.91 & 69.46 & 21.48 & 32.91 & 71.43 & 47.88 \\
9 & 60.79 & 69.26 & 74.38 & 71.13 & 21.48 & 32.91 & 68.77 & 45.62 \\
10 & 59.51 & 68.37 & 74.98 & 71.63 & 21.48 & 25.91 & 69.75 & 52.91 \\
11 & 55.27 & 70.05 & 72.61 & 60.99 & 21.48 & 30.94 & 68.77 & 49.75 \\
12 & 59.70 & 71.23 & 71.63 & 63.94 & 21.48 & 40.30 & 72.12 & 65.02 \\
13 & 45.81 & 67.88 & 73.10 & 64.83 & 21.48 & 37.04 & 69.46 & 55.17 \\
14 & 42.17 & 67.98 & 73.30 & 62.56 & 21.48 & 40.49 & 71.92 & 67.00 \\
15 & 33.79 & 60.30 & 71.13 & 61.67 & 21.48 & 37.24 & 73.79 & 66.80 \\
16 & 40.59 & 61.38 & 67.98 & 51.43 & 21.48 & 36.16 & 73.60 & 67.68 \\
17 & 36.26 & 65.42 & 69.16 & 52.71 & 21.48 & 36.75 & 73.69 & 58.33 \\
18 & 45.12 & 67.29 & 70.54 & 47.88 & 21.48 & 35.96 & 73.00 & 61.38 \\
19 & 40.00 & 69.06 & 70.05 & 41.87 & 21.58 & 33.30 & 73.10 & 58.82 \\
20 & 28.08 & 56.95 & 71.92 & 64.14 & 21.48 & 32.71 & 72.02 & 44.53 \\
21 & 41.38 & 55.47 & 77.04 & 73.99 & 21.48 & 37.04 & 75.47 & 42.17 \\
22 & 38.62 & 63.74 & 76.35 & 74.09 & 21.48 & 37.93 & 77.24 & 57.24 \\
23 & 43.05 & 66.11 & 75.47 & 76.55 & 21.87 & 38.72 & 77.44 & 63.35 \\
24 & 38.13 & 60.89 & 76.55 & 76.45 & 25.42 & 39.90 & 79.70 & 61.97 \\
25 & 55.96 & 68.08 & 77.14 & 79.11 & 40.69 & 54.48 & 71.13 & 66.80 \\
26 & 71.72 & 74.09 & 76.65 & 79.41 & 69.75 & 71.13 & 79.01 & 74.19 \\
27 & 71.72 & 73.99 & 75.96 & 78.72 & 75.86 & 72.71 & 79.11 & 79.41 \\
28 & 70.74 & 73.50 & 75.86 & 78.62 & 73.10 & 72.71 & 78.23 & 80.20 \\
29 & 73.60 & 74.38 & 75.57 & 78.03 & 77.83 & 76.45 & 76.75 & 79.41 \\
30 & 72.22 & 73.89 & 75.67 & 77.14 & 74.68 & 74.58 & 77.14 & 80.30 \\
31 & 71.63 & 73.50 & 75.67 & 77.34 & 73.40 & 73.99 & 77.83 & 80.10 \\
32 & 71.53 & 73.50 & 75.67 & 77.14 & 74.38 & 74.68 & 75.07 & 76.75 \\
33 & 72.91 & 73.89 & 75.07 & 77.54 & 72.81 & 73.89 & 74.19 & 76.16 \\
34 & 72.32 & 74.09 & 75.57 & 77.34 & 74.38 & 74.88 & 74.78 & 75.96 \\
35 & 71.23 & 73.60 & 75.37 & 76.35 & 69.75 & 72.61 & 75.17 & 76.95 \\
\midrule
\multicolumn{9}{c}{\textbf{Qwen-2.5-7B}} \\
\midrule
1  & 85.52 & 85.71 & 86.21 & 85.42 & 29.66 & 67.19 & 81.28 & 29.85 \\
2  & 85.81 & 86.11 & 85.81 & 85.12 & 37.93 & 67.78 & 79.90 & 35.76 \\
3  & 86.21 & 85.91 & 85.71 & 85.02 & 33.20 & 78.13 & 82.76 & 62.17 \\
4  & 86.31 & 86.11 & 84.33 & 83.35 & 44.04 & 80.79 & 84.43 & 73.89 \\
5  & 85.62 & 85.62 & 84.93 & 83.94 & 59.51 & 83.74 & 83.98 & 77.34 \\
6  & 85.22 & 85.42 & 85.52 & 84.24 & 58.42 & 82.76 & 82.56 & 73.69 \\
7  & 84.33 & 85.12 & 84.53 & 82.17 & 54.58 & 81.67 & 83.94 & 74.98 \\
8  & 81.18 & 84.63 & 85.32 & 82.56 & 38.82 & 66.40 & 83.94 & 74.19 \\
9  & 83.25 & 84.33 & 85.71 & 82.27 & 53.69 & 79.80 & 85.12 & 76.45 \\
10 & 73.89 & 84.14 & 84.04 & 80.49 & 54.78 & 80.10 & 85.02 & 71.23 \\
11 & 73.69 & 83.35 & 82.17 & 77.93 & 57.34 & 81.48 & 84.73 & 75.07 \\
12 & 70.25 & 81.38 & 82.56 & 82.17 & 57.14 & 80.39 & 85.42 & 70.64 \\
13 & 26.40 & 72.12 & 76.45 & 63.25 & 47.39 & 72.71 & 84.73 & 65.62 \\
14 & 24.53 & 71.82 & 77.44 & 61.38 & 21.87 & 76.45 & 83.55 & 39.80 \\
15 & 25.62 & 61.38 & 77.83 & 22.76 & 22.86 & 68.18 & 85.71 & 45.62 \\
16 & 21.97 & 58.23 & 76.95 & 23.84 & 21.48 & 57.44 & 80.30 & 23.94 \\
17 & 39.70 & 73.50 & 74.09 & 23.35 & 22.86 & 25.81 & 60.00 & 24.43 \\
18 & 56.65 & 83.25 & 84.63 & 77.24 & 55.96 & 70.25 & 86.31 & 59.31 \\
19 & 76.85 & 83.25 & 86.50 & 84.53 & 45.42 & 77.14 & 86.50 & 52.32 \\
20 & 84.73 & 85.81 & 85.42 & 84.63 & 82.86 & 83.94 & 87.00 & 61.28 \\
21 & 84.63 & 85.81 & 85.62 & 84.93 & 83.84 & 85.32 & 86.70 & 48.08 \\
22 & 86.01 & 86.01 & 84.53 & 83.94 & 83.55 & 85.81 & 86.50 & 83.65 \\
23 & 86.50 & 86.31 & 84.73 & 84.24 & 68.57 & 85.42 & 86.50 & 86.70 \\
24 & 86.11 & 86.11 & 85.42 & 85.02 & 69.66 & 86.21 & 84.73 & 83.45 \\
25 & 86.60 & 85.62 & 85.42 & 84.83 & 75.17 & 86.70 & 84.43 & 81.58 \\
26 & 83.84 & 84.93 & 85.71 & 85.52 & 78.23 & 83.65 & 85.42 & 84.43 \\
\bottomrule
\end{tabular*}
\caption{{Detailed MedQA-CH accuracy under mean-difference activation steering.
\textsc{Dual} holds the answer option fixed and contrasts the \textsc{Sup} and \textsc{Elim} instructions. CAA holds the instruction fixed and contrasts the correct and paired incorrect options. Results use steering coefficients $a \in \{-2,-1,1,2\}$.
The upper block reports GLM-4-9B results, while the lower block reports Qwen-2.5-7B results.
The unsteered case $a=0$ is omitted for clarity.
}}
\label{tab:full_steering_results}
\end{table*}

\section{Representation and Window Analyses}
\label{app:additional_layer_analysis}

\subsection{Additional Visualization of Layer-wise Separability}
\label{app:pca_vis}

We visualize \texttt{[STATE]} representations using principal component analysis (PCA) to illustrate their layer-wise geometry under {the two framings}.

For Qwen-2.5-7B, we extract residual-stream activations at the \texttt{[STATE]} position using the same contiguous feature window ($w=6$) as in the separability analysis. At each layer, PCA is applied independently to pooled representations from {both framings}, and the first three principal components are visualized.

\begin{figure*}[p]
\centering
    \includegraphics[height=0.72\textheight,keepaspectratio]{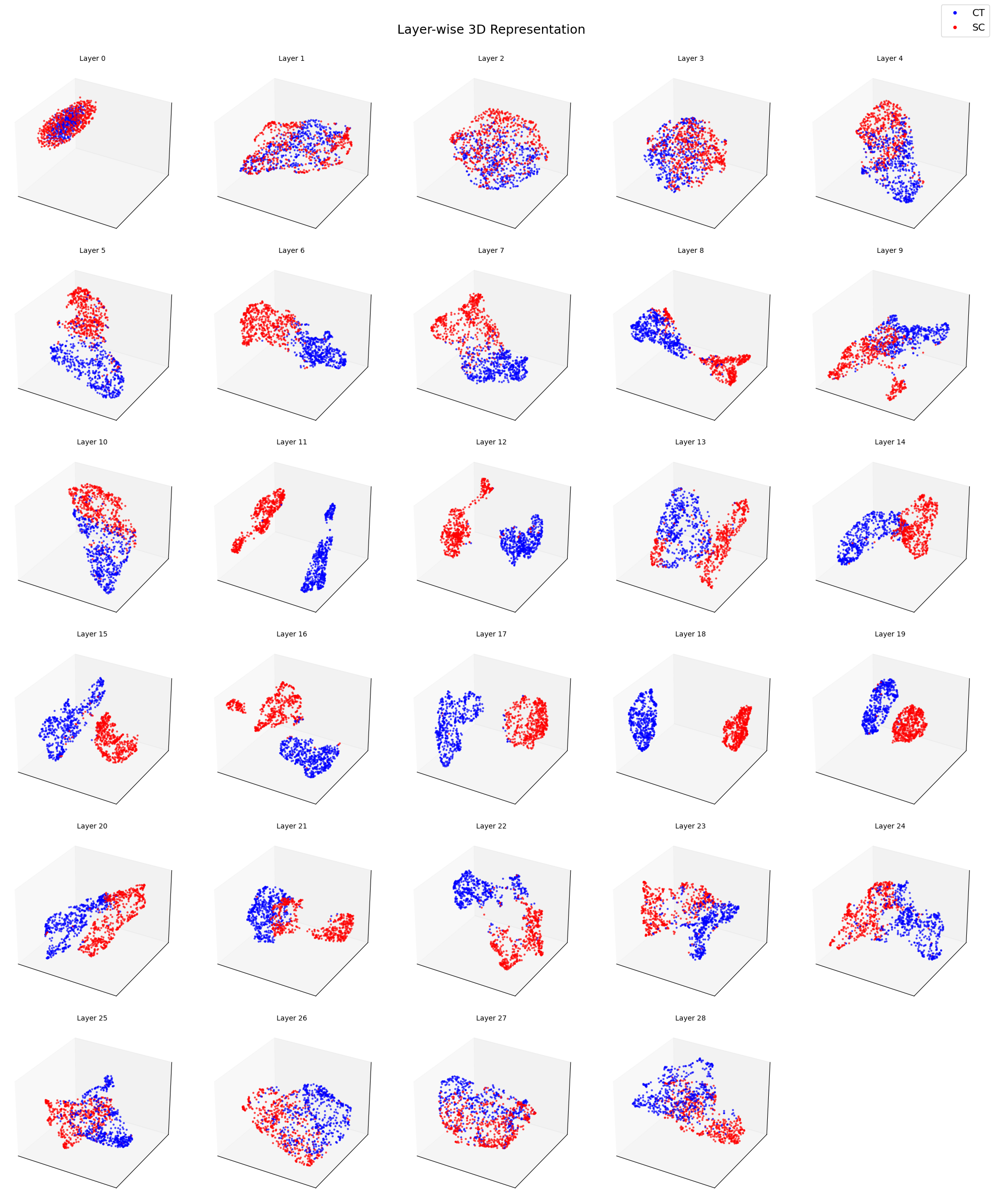}
\caption{\textbf{Layer-wise PCA of {framing-conditioned} \texttt{[STATE]} representations.}
For each layer of Qwen-2.5-7B, PCA is applied to pooled representations using the same feature window ($w=6$) as in the separability analysis. Points are individual instances colored by {framing}. Intermediate-layer representations show clearer separation than early- or late-layer representations, consistent with Figure~\ref{fig:cross_model_cohen_d_heatline}.}
\label{fig:pca_layers}
\end{figure*}

Figure~\ref{fig:pca_layers} shows substantial overlap in early layers, clearer separation in intermediate layers, and reduced separation in later layers, matching the non-monotonic depth-wise trend in Figure~\ref{fig:cross_model_cohen_d_heatline}.

PCA is performed independently at each layer; hence the resulting axes are not comparable across layers.

\subsection{Analysis of the Intervention Window}
\label{app:window_analysis}
This subsection compares candidate layer windows using output-level statistics under representation substitution as a post-hoc assessment of the separability-selected intervention window.

\paragraph{Analysis Metrics}
We measure (i) cosine distance between the original and intervened logit vectors (global geometric change), and
(ii) $\mathrm{KL}(q\|p)$ between the corresponding output distributions (probability mass reallocation).
For each metric we report the mean and two stability summaries: IQR and the 90th percentile (p90), where larger p90 indicates heavier tails. p90 corresponds to the 90th percentile of per-example distances; larger p90 implies more extreme cases among the top 10\% most affected examples.

\begin{table}[t]
\centering
\small
\begin{tabular}{lccc|ccc}
\toprule
\textbf{Layers} 
& \textbf{Cos} $\downarrow$ & \textbf{IQR} & \textbf{p90}
& \textbf{KL} $\uparrow$ & \textbf{IQR} & \textbf{p90} \\
\midrule
3--5     & 0.130 & 0.070 & 0.275 & 1.634 & 1.319 & 6.832 \\
13--15   & 0.108 & 0.085 & 0.295 & 1.000 & 0.943 & 3.745 \\
11--20   & 0.108 & 0.066 & 0.257 & 1.487 & 1.198 & 6.253 \\
25--28   & 0.131 & 0.069 & 0.275 & 1.636 & 1.317 & 6.832 \\
\bottomrule
\end{tabular}
\caption{Window comparison under representation substitution. Lower cosine distance indicates smaller global distortion in logit space; when paired with non-trivial KL shifts, we use it as a proxy for reduced collateral perturbation. IQR and p90 summarize dispersion and tail heaviness. Values are averaged across swap directions.}
\label{tab:window_analysis}
\end{table}

\paragraph{Layer Windows Discussion}
Table~\ref{tab:window_analysis} aggregates results across swap directions (direction-specific trends are qualitatively consistent with the conclusions).
Shallow (3--5) and late (25--28) layers behave nearly identically on both metrics, yielding larger cosine distances (0.130--0.131) and the largest KL values (1.634--1.636) with heavier KL tails (p90=6.832). 
This suggests strong distributional shifts accompanied by more extreme probability reallocations among the most affected examples.

The intermediate-layer peak (13--15) exhibits the largest dispersion and tail heaviness in cosine distance (IQR=0.085, p90=0.295) while producing the smallest KL magnitude and tails (KL mean=1.000, p90=3.745), indicating a mismatch between geometric logit changes and distributional reallocation under this window.

In contrast, the intermediate window (11--20) shows the most favorable stability--effect profile under our criteria: mean cosine distance is similar for 11--20 and 13--15 (both $\approx 0.108$), but 11--20 achieves the smallest cosine dispersion and lightest cosine tails (IQR=0.066, p90=0.257), while maintaining substantial KL shifts (mean=1.487) and a lighter KL tail compared to 3--5/25--28 (p90 6.253 vs.\ 6.832). 
A smaller KL p90 implies less extreme probability reallocations among the top 10\% most affected instances, which is desirable for a robust and predictable intervention interface.

\paragraph{Conclusion}
Overall, the separability-selected window 11--20 also provides the best stability--effect trade-off under output-level statistics: it avoids large and heavy-tailed geometric perturbations (cosine) while still inducing meaningful distributional changes (KL) without the heaviest tails observed at the layer-range extremes. This analysis is post hoc: the output-level statistics are not used to select the intervention window or tune it for task performance.

\subsection{Cross-Model Separability Diagnostics at the \texttt{[STATE]} Interface}
\label{app:cross_model_separability}

To complement the PCA visualization in Section~\ref{sec:exp_internal}, we report the full layer-wise separability diagnostics for both Qwen-2.5-7B and GLM-4-9B. For each model, we use the same paired Sup/Elim construction, compute Cohen's $d$ over residual-stream features at the \texttt{[STATE]} position, and apply the same sliding-window procedure over feature dimensions. We report both the label-free and label-informed variants of the normalized separability heatmaps.

\begin{figure}[H]
\centering
\includegraphics[width=\linewidth]{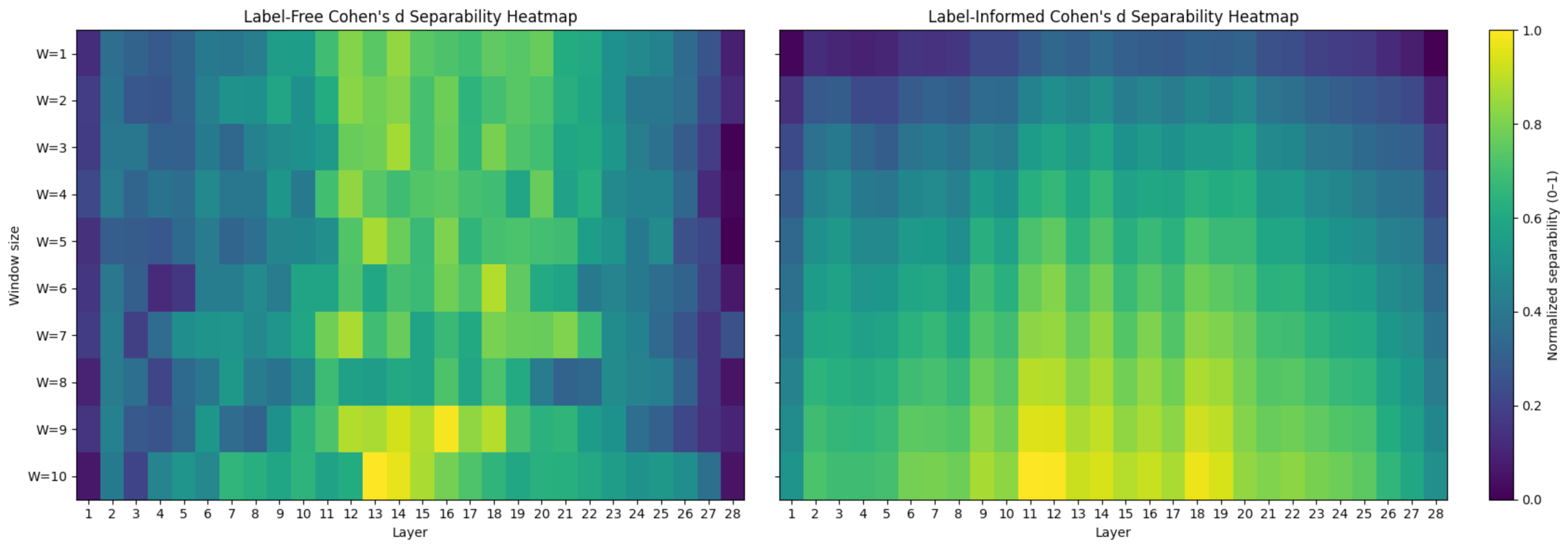}
\vspace{1mm}

\includegraphics[width=\linewidth]{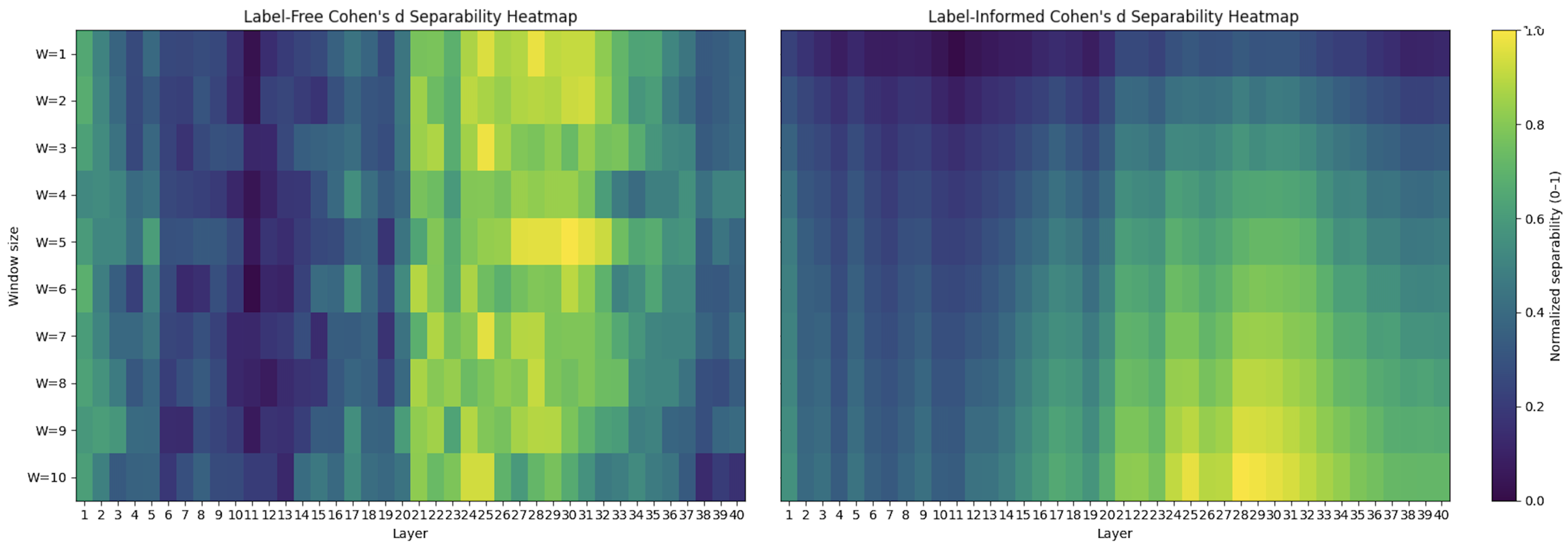}
\caption{
Layer-wise normalized separability at the \texttt{[STATE]} position for Qwen-2.5-7B (top) and GLM-4-9B (bottom).
For each model, the heatmaps report label-free Cohen's $d$ and the corresponding label-informed variant.
Both models show weak early-layer separability, a pronounced intermediate-layer concentration, and attenuation toward later layers.
}
\label{fig:cross_model_cohen_d_heatline}
\end{figure}

Figure~\ref{fig:cross_model_cohen_d_heatline} shows that the two models exhibit a consistent layer-wise pattern. In both Qwen-2.5-7B and GLM-4-9B, separability is weak in early layers, becomes most pronounced in intermediate layers, and attenuates toward later layers. This cross-model consistency supports the conclusion that framing-dependent structure at the \texttt{[STATE]} interface is concentrated in an intermediate-layer region rather than uniformly distributed across network depth.

Following the same criterion used in the main text, we select model-specific intervention windows based solely on the observed separability patterns: layers 11--20 for Qwen-2.5-7B and layers 21--40 for GLM-4-9B. These windows are not tuned based on downstream task performance; they are used in subsequent experiments to test whether \texttt{[STATE]} substitution within the separable intermediate-layer region produces systematic behavioral changes.

\subsection{Robustness of Window Localization}
\label{app:window_robustness}

{Window localization uses a calibration set of 50 examples. Each example is instantiated with two prompt realizations for each of the \textsc{Sup} and \textsc{Elim} framings, yielding 200 framing-conditioned representations. The resulting model-specific intervention window is fixed before evaluation on the test split.}

{We assess the stability of this localization for Qwen-2.5-7B on an expanded sample of 1,000 examples. We generate 500 bootstrap resamples and recompute the layer-wise separability profile for each resample. Table~\ref{tab:window_bootstrap} reports the mean normalized score and the width of the 95\% bootstrap confidence interval for the ten highest-scoring layers under each diagnostic.}

\begin{table}[t]
\centering
\resizebox{\columnwidth}{!}{%
\begin{tabular}{rccc@{\hspace{4mm}}rccc}
\toprule
\multicolumn{4}{c}{\textbf{Label-free separability}} &
\multicolumn{4}{c}{\textbf{Label-informed separability}} \\
\cmidrule(lr){1-4}\cmidrule(lr){5-8}
\textbf{Rank} & \textbf{Layer} & \textbf{Mean} & \makecell{\textbf{95\% CI}\\\textbf{width}} &
\textbf{Rank} & \textbf{Layer} & \textbf{Mean} & \makecell{\textbf{95\% CI}\\\textbf{width}} \\
\midrule
1  & 19 & 0.898 & 0.142 & 1  & 12 & 0.730 & 0.205 \\
2  & 16 & 0.891 & 0.139 & 2  & 14 & 0.716 & 0.181 \\
3  & 14 & 0.882 & 0.137 & 3  & 11 & 0.705 & 0.152 \\
4  & 20 & 0.868 & 0.142 & 4  & 18 & 0.705 & 0.200 \\
5  & 18 & 0.853 & 0.139 & 5  & 19 & 0.682 & 0.178 \\
6  & 12 & 0.839 & 0.135 & 6  & 16 & 0.674 & 0.175 \\
7  & 13 & 0.807 & 0.135 & 7  & 13 & 0.665 & 0.186 \\
8  & 9  & 0.775 & 0.145 & 8  & 20 & 0.660 & 0.180 \\
9  & 11 & 0.767 & 0.124 & 9  & 15 & 0.636 & 0.173 \\
10 & 17 & 0.747 & 0.123 & 10 & 17 & 0.635 & 0.174 \\
\bottomrule
\end{tabular}
}
\caption{{Bootstrap summaries of layer-wise separability for Qwen-2.5-7B. For each diagnostic, the table lists the ten layers with the highest mean normalized scores; confidence-interval widths are estimated from 500 bootstrap resamples of the 1,000-example sample.}}
\label{tab:window_bootstrap}
\end{table}

{Under both diagnostics, the highest-scoring layers cluster in the intermediate region, with most falling between Layers 11 and 20. The bootstrap results show that localization to this region is consistent across resamples, although the exact layer rankings differ between the two diagnostics.}

\section{Ablation Details and Controls}
\label{app:ablation_details}

This section provides implementation details for the ablation experiments described in the main text.
All interventions operate on the residual-stream activations at the \texttt{[STATE]} position and are applied layerwise within the intervention window $W$, unless otherwise specified.

\subsection{Token and Interface Controls}
\label{app:token_interface_controls}

\begin{table}[H]
\centering
\small
\begin{tabular}{lcc}
\toprule
\textbf{Setting} & \textbf{Correct / Total} & \textbf{Accuracy} \\
\midrule
GLM-4-9B w/o token & 812 / 1015 & 80.00\% \\
With \texttt{[STATE]} token & 810 / 1015 & 79.80\% \\
\midrule
Qwen-2.5-7B w/o token & 882 / 1015 & 86.90\% \\
With \texttt{[STATE]} token & 885 / 1015 & 87.19\% \\
\bottomrule
\end{tabular}
\caption{Latent option-preference accuracy with and without the \texttt{[STATE]} token.}
\label{tab:state_token_ablation}
\end{table}

To test interface specificity, we compare substitution at \texttt{[STATE]} with otherwise identical substitution at the original prompt's final content token. Substitution content and the layer window are held fixed in the interface comparison.

Accuracies in Table ~\ref{tab:state_token_ablation} are obtained by ranking the four option tokens {A, B, C, D} at the final layer under a constrained scoring protocol, and therefore index the model's latent preference over answer options rather than its generated answer. They are not comparable to the CoT-parsed accuracies in Table 1, which are produced by autoregressive generation followed by option-set parsing under a different prompt structure. This control therefore establishes that inserting [STATE] leaves the latent option distribution essentially unchanged; it does not by itself rule out effects of token insertion that accumulate over long-form generation, and the BASE condition in Table 1 already includes the [STATE]-augmented prompt so that insertion effects cancel within the BASE→SUB contrast.

\subsection{Substitution content}
\label{app:substitution_content}
We consider the following substitution types when modifying the \texttt{[STATE]} representation:

\begin{itemize}
    \item \textbf{Zeroing.}
    The residual vector at the \texttt{[STATE]} position is replaced with a zero vector at each layer in $W$.

    \item \textbf{Random replacement.}
    The residual vector is replaced by a randomly sampled vector drawn from the empirical distribution of \texttt{[STATE]} activations at the same layer.
    Sampling is performed independently for each layer, and the sampled vector is fixed across forward passes within a run.

    \item \textbf{{Framing-conditioned} substitution.}
    The residual vector at the \texttt{[STATE]} position is replaced with a cached activation obtained from {the complementary framing} (Sup$\leftrightarrow$Elim) for the same input question.
    Substitution is applied layerwise within $W$, preserving the original activation outside the intervention window.
\end{itemize}

\subsection{Interface Position}
\label{app:sub_position}

To test whether substitution depends on the structural role of the interface, we compare the canonical final-position configuration with two misaligned placements:
\[
\mathbf{x}_1 = [\texttt{PAD}]^{n} + I + Q + O + \texttt{[STATE]},
\]
\[
\mathbf{x}_2 = [\texttt{PAD}]^{n} + I + Q + \texttt{[STATE]} + O,
\]
\[
\mathbf{x}_3 = [\texttt{PAD}]^{n} + I + \texttt{[STATE]} + Q + O,
\]
where $I$, $Q$, and $O$ denote the instruction, question, and answer-option segments, respectively. For all three configurations, we apply the same {framing-conditioned} substitution over the same layer window $W$.

We count a generation as failed when it is incoherent or cannot be parsed into the required task-level answer representation. Across the two misaligned configurations, more than 95\% of generations fail this criterion. By contrast, the canonical configuration preserves coherent, task-relevant generation for the majority of inputs. This result indicates that \texttt{[STATE]} is not an arbitrary writable slot: its effectiveness depends on being placed after the complete question and option context.

\subsection{Quantitative Similarity Summaries for Ablations}
\label{app:quant_summaries}

This appendix reports distributional summaries for the response-level similarity analyses shown in Figures~\ref{fig:dist_a} and~\ref{fig:dist_b}.
For each condition, we compute length-aware cosine similarity (LAC) and BERTScore-F1 between the baseline output and the output under intervention, and report the median, interquartile range (IQR), and upper quantiles (P90/P95). The number of evaluated runs is denoted by $N$.

\paragraph{Substitution content: structured vs.\ unstructured interventions}
Table~\ref{tab:quant_subst_content} summarizes the similarity distributions for different substitution contents at the \texttt{[STATE]} position (Figure~\ref{fig:dist_b}), contrasting {framing-conditioned} substitution with content-agnostic controls.

\begin{table}[t]
\centering
\small
\begin{tabular}{lcccc}
\toprule
\textbf{Metric} & \textbf{Median} & \textbf{IQR} & \textbf{P90} & \textbf{P95} \\
\midrule
\multicolumn{5}{l}{\textbf{LAC}} \\
\makecell[l]{\quad zeroing}                 & 0.6620 & 0.4913 & 0.9005 & 0.9351 \\
\makecell[l]{\quad random}      & 0.6242 & 0.4933 & 0.8931 & 0.9141 \\
\makecell[l]{\quad {cross-framing}} & 0.8766 & 0.1925 & 1.0000 & 1.0000 \\
\midrule
\multicolumn{5}{l}{\textbf{BERTScore-F1}} \\
\makecell[l]{\quad zeroing}                 & 0.5631 & 0.1196 & 0.6954 & 0.7554 \\
\makecell[l]{\quad random}      & 0.5560 & 0.1289 & 0.6855 & 0.7460 \\
\makecell[l]{\quad {cross-framing}} & 0.6400 & 0.3449 & 1.0000 & 1.0000 \\
\bottomrule
\end{tabular}
\caption{Quantile summaries of response-level similarity for different substitution contents at the \texttt{[STATE]} position (Figure~\ref{fig:dist_b}).}
\label{tab:quant_subst_content}
\end{table}

\paragraph{Intervention location: \texttt{[STATE]} vs.\ final-content-token substitution}
Table~\ref{tab:quant_lasttok} reports quantile summaries for substitutions applied at \texttt{[STATE]} versus at the final content token position (Figure~\ref{fig:dist_a}), using the same intervention window $W$ and decoding procedure.

\begin{table}[t]
\centering
\small
\begin{tabular}{lcccc}
\toprule
\textbf{Metric} & \textbf{Median} & \textbf{IQR} & \textbf{P90} & \textbf{P95} \\
\midrule
\multicolumn{5}{l}{\textbf{LAC}} \\
\makecell[l]{\quad [STATE]} & 0.8766 & 0.1925 & 1.0000 & 1.0000 \\
\makecell[l]{\quad Final content\\\quad token} & 0.9780 & 0.0660 & 1.0000 & 1.0000 \\
\midrule
\multicolumn{5}{l}{\textbf{BERTScore-F1}} \\
\makecell[l]{\quad [STATE]} & 0.6400 & 0.3449 & 1.0000 & 1.0000 \\
\makecell[l]{\quad Final content\\\quad token} & 0.9444 & 0.2330 & 1.0000 & 1.0000 \\
\bottomrule
\end{tabular}
\caption{Quantile summaries comparing substitutions at the \texttt{[STATE]} position and the final content token.}
\label{tab:quant_lasttok}
\end{table}

\end{document}